\documentclass[acmtog]{acmart}

\usepackage{booktabs} 
\usepackage{hyperref}
\usepackage{url}
\usepackage{amsmath}
\usepackage{bm}
\usepackage{graphicx}
\usepackage{wrapfig}
\usepackage[noend]{algorithmic}
\usepackage{float}
\usepackage{subcaption}
\usepackage{multirow}
\newcommand{\citewithauthor}[1]{\citeauthor{#1} \shortcite{#1}}
\usepackage{pifont}
\usepackage[ruled]{algorithm2e} 

\SetAlFnt{\small}
\SetAlCapFnt{\small}
\SetAlCapNameFnt{\small}
\SetAlCapHSkip{0pt}

\acmJournal{TOG}

\definecolor{green}{rgb}{0, 0.5, 0}
\definecolor{orange}{rgb}{1.0, 0.6, 0.2}
\definecolor{red}{rgb}{1.0, 0.0, 0.0}
\definecolor{blue}{rgb}{0.0, 0.0, 1.0}
\definecolor{teal}{rgb}{0.0, 0.4, 0.4}
\definecolor{purple}{rgb}{0.65,0,0.65}
\definecolor{saffron}{rgb}{0.95,0.75,0.2}
\definecolor{turquoise}{rgb}{0.0,0.5,0.5}
\definecolor{black}{rgb}{0,0,0}

\newcommand{\zhf}[1]{{\color{black}#1}}
\newcommand{\zhn}[1]{{\color{black}#1}}

\usepackage{amsmath}
\DeclareMathOperator*{\argmax}{arg\,max}

\begin{document}
\title{Learning Physically Realizable Skills for Online Packing\\of General 3D Shapes}


\author{Hang Zhao}
\affiliation{%
 \institution{National University of Defense Technology and Nanjing University}
 \country{China}
 }
\email{alex.hang.zhao@gmail.com}

\author{Zherong Pan}
\affiliation{%
 \institution{Lightspeed Studio, Tencent America}
 \country{USA}
}
\email{zherong.pan.usa@gmail.com}

\author{Yang Yu}
\affiliation{%
\institution{Nanjing University}
\country{China}
\department{National Key Laboratory for Novel Software Technology}
}
\email{yuy@lamda.nju.edu.cn}

\author{Kai Xu}
\authornote{Kai Xu is the corresponding author.}
\affiliation{%
 \institution{National University of Defense Technology}
 \country{China}
}
\email{kevin.kai.xu@gmail.com}

\begin{teaserfigure}
\centering
\includegraphics[width=1\textwidth]{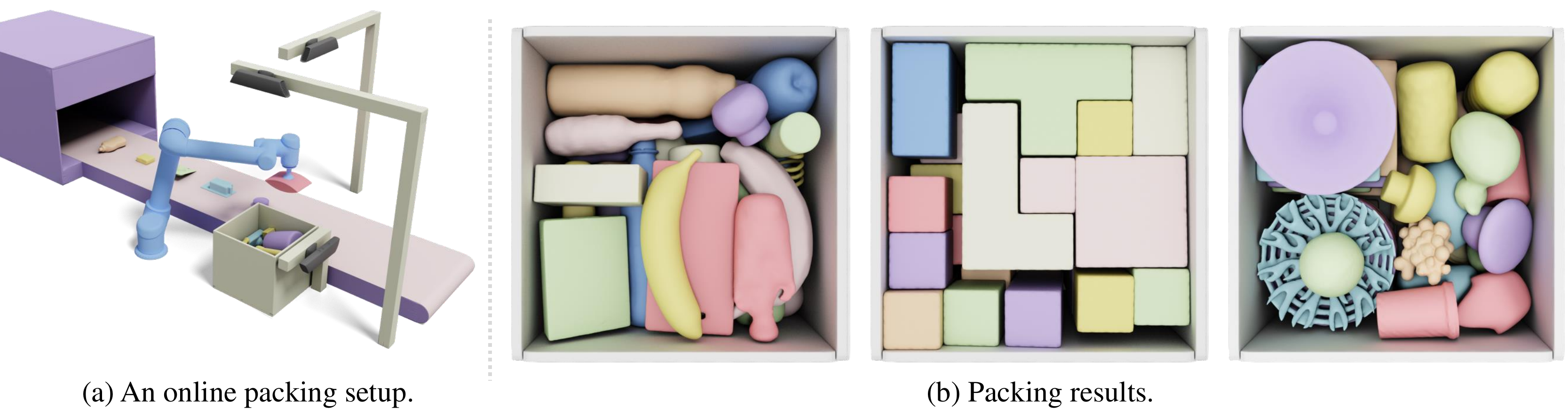}
\caption{\label{fig:teaser} We develop a learning-based solver for packing arbitrarily-shaped objects in a  physically realizable problem setting. This figure shows an online virtual packing setup (a), where objects move on a conveyor belt at a constant speed.
Three RGB-D cameras are provided for observing the top and bottom surfaces of the incoming object as well as  packed object configurations inside the container.
Within a limited time window, the robot has to decide on the placement for the incoming object in the target container for tight packing. Packing results on 3D shapes with different geometric properties are shown in (b).}
\vspace{10pt}
\end{teaserfigure}
\begin{abstract}
We study the problem of learning \emph{online} packing skills for \emph{irregular 3D shapes}, \zhn{which is arguably the most challenging setting of bin packing problems.}
The goal is to consecutively move a sequence of 3D objects with arbitrary shapes into a designated container with only partial observations of the object sequence.
Meanwhile, we take physical realizability into account, involving physics dynamics 
and constraints of a placement. The packing policy should understand the 3D geometry of the object to be packed and make effective decisions \zhn{to accommodate it in the container in a physically realizable way}.
We propose a Reinforcement Learning (RL) pipeline to learn the policy.
The complex irregular geometry and imperfect object placement together lead to huge solution space. Direct training in such space is prohibitively
data intensive. We instead propose a theoretically-provable method for candidate action generation to reduce the action space of RL \zhn{and the learning burden}.
A parameterized policy is then learned to select the best placement from the candidates. Equipped with an efficient method of asynchronous RL acceleration and a data preparation process of simulation-ready training sequences, a mature packing policy can be trained in a physics-based environment within 48 hours. 
Through extensive evaluation on a variety of real-life shape datasets \zhn{and comparisons with  state-of-the-art baselines, we demonstrate that} our method outperforms the best-performing baseline on all datasets by at least $12.8\%$ in terms of packing utility. We also release our datasets and source code to support further research in this direction \footnote{Datasets and source code available at \url{https://github.com/alexfrom0815/IR-BPP}.}.

\end{abstract}


%
%

\maketitle
\section{\label{sec:intro}Introduction}
Packing, finding an efficient placement of as many as possible objects within a designated volume, has garnered multidisciplinary research interests from combinatorial optimization~\citep{MartelloPV00, Seiden02}, computational geometry~\citep{MaCHW18,HuXCG0020} and machine learning~\cite{zhao2022learning, hu2017solving}. 
The famous Kepler conjecture, which considers the continuous packing of spheres in infinite spaces, was proved only recently~\cite{hales2017formal} after more than four centuries.
Even the discrete bin packing problem has been proven NP-hard~\cite{hartmanis1982computers}. For 3D shape packing, most existing works consider simple object shapes such as cuboids~\cite{MartelloPV00}, tetrahedra~\cite{conway2006packing}, or ellipsoids~\cite{Kallrath17}. The more general problem of \emph{irregular shape packing}, on the other hand, has received much less study, although being practically useful in many real application scenarios. In robotics, product packing robots~\cite{WangH19a,8793966,wang2020robot,yang2021packerbot} for logistics automation is an active research area. In computer graphics, irregular shape packing has been widely explored in UV atlas generation~\cite{LimperVS18,Liu_AAAtlas_2019,ZHANG2020101854}, artistic puzzle design~\cite{wang2021mocca,Chen-2022-HighLevelPuzzle}, 2D panel fabrication~\cite{SaakesCMI13}, and 3D printing~\cite{ChenZLHLHBCC15}, with various constraints.


We focus on  
practically feasible robotic packing, namely \emph{Physically Realizable Packing (PRP)}. In particular, PRP combines two well-known robotic tasks, pick-and-place~\cite{han2019toward} and irregular shape packing~\cite{WangH19a}, \zhn{while further requiring packed objects governed by physics dynamics.}
As illustrated in~Fig.~\ref{fig:teaser}, our virtual problem setup involves a robot arm equipped with a sucker-type gripper. Irregularly shaped objects are transported by a conveyor belt in a planar-stable pose and move at a constant speed.
The upper surface of the object is captured by a top-view camera and the robot can move this object above an up-looking camera to capture its bottom surface.
We pursue an online setting ~\citep{Seiden02} 
where the robot observes only the object coming in the next instead of the full sequence.
After the robot releases each object 
at its planned configuration inside the target container, 
we use a full-fledged physics simulator~\cite{coumans2016pybullet}, compatible with the standard Reinforcement Learning (RL) platform~\cite{BrockmanCPSSTZ16}, to determine the ultimate quasi-static poses of all objects and enforce physically realizable constraints. 
Our physics constraint accounts for 
both quasi-static and dynamic interactions between objects, generalizing the prior pile stability constraint~\cite{WangH19a} which only considers quasi-static interactions.

We propose a novel RL pipeline to train effective packing policies. The sequential nature of packing has stimulated several recent RL-based approaches~\cite{HuXCG0020,ZhaoS0Y021,zhao2022learning}. Compared with manually designed heuristics~\cite{KarabulutI04,ramos2016container, ha2017online}, RL is capable of learning complex application-side constraints from 
guided explorations. RL also bears the potential to outperform humans on both continuous and discrete decision-making problems~\cite{MnihKSRVBGRFOPB15,duan2016benchmarking}. However, prior RL-based approaches opt to factor out the continuous aspect of the problem and only learn a discrete packing policy, via assuming cuboid objects and omitting physics constraints. The continuous nature of the irregular shape packing calls for exploiting the full potential of RL through accounting for physics constraints.
We contribute a practical algorithm to learn packing  policies for  irregular 3D shapes via overcoming a series of technical challenges.

First of all, learning effective packing  policies is naturally a tough challenge.
The complex irregular geometry and imperfect object placement \zhn{due to physics dynamics} together lead to huge solution space. Direct policy training through trial and error in such spaces  is prohibitively data intensive.
We instead propose a candidate action generation method to reduce the action space of RL \zhn{as well as the learning burden}. The candidates are the convex vertices of the (polygonal) connected regions within which the current object can be placed. We prove that these candidates are local optima that make the object tightly packed against the obstacles (the placed objects and the container). Our learned policy is then used to understand the geometry of the current object and choose the best placement from its corresponding candidates.
This method significantly reduces the search space of RL and enables reliable learning of packing policies.

Both the geometry understanding and the candidate selection need a gazillion experiences 
collected through simulation. 
However, interaction with the simulation world is CPU-bound and is time-consuming, which leaves policies less trained per wall-clock time.
Some abnormally slow instances also block the uniform training schedule.
To this end, we propose to accelerate the training via an asynchronous sampling strategy.
In particular, we decouple the training into a parallelized experience sampling process for non-blocking treatment and a separate learning process for continuously updating parameters. 
Our method allows a robust packing policy to be trained within 48 hours on a desktop machine.

In addition, RL algorithms should be trained with sufficient data variety to ensure the robustness of the learned policies. \zhf{As compared with cubical shape packing, however, the variety of object shapes and poses for irregular shape packing is on a much higher level.} We propose a data preparation process for generating sequences of simulation-ready objects each with a planar-stable pose. Working with a combination of several real-world and synthetic datasets, we create a packing problem generator emitting randomized, versatile, and faithful problem instances.


We have conducted extensive evaluations of our method on well-established datasets with various geometric characteristics. By comparing with a row of baseline algorithms, \zhn{we demonstrate that our method significantly outperforms the best-performing baseline on all datasets by at least $12.8\%$ in terms of packing utility.}
Furthermore, we extend our method to the scenario of buffered packing, where the robot maintains \zhf{a buffer} for re-ordering objects, and show that higher packing quality can be achieved.
Our contributions include:
\begin{itemize}
\item An effective and theoretically-provable placement candidate generation method for pruning the action space of RL, along with a learnable packing policy for candidate selection.
\item An efficient, asynchronous, off-policy 
sampling strategy
for accelerating packing policy training.
\item A constructive PRP environment modeling realistic packing with a large object dataset and RL-compatible interfaces.
\end{itemize}

\section{\label{sec:related}Related Work}
Our work is built off of prior works on packing policy searches. Our problem is also closely related to other packing-related tasks in computer graphics and robotics.

\subsection{Packing Policies} 
\begin{figure}[!t]
\centering
\includegraphics[width=0.5\textwidth]{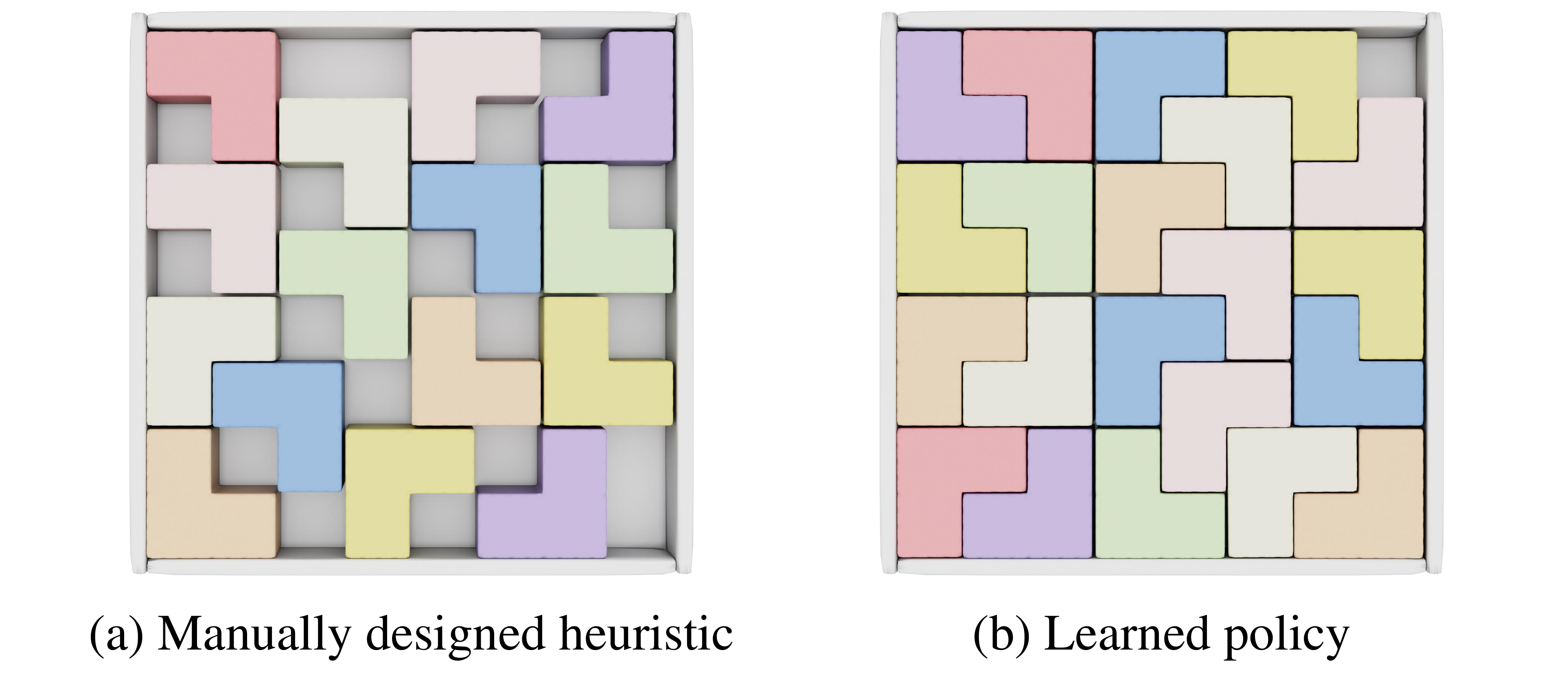}
\vspace{-15pt}
\caption{\label{fig:Insight} When packing simple polyominos in a flat container, the result generated using the HM heuristic (a) exhibits a non-trivial sub-optimality gap as compared with that of our learned policy (b).}
\vspace{-10pt}
\end{figure}

Being an NP-hard problem, various heuristic policies have been proposed for 2D~\cite{lodi2002two} and 3D~\cite{ali2022line} cubical object packing over time, without a single best performer. The transition of focus to irregular shapes is quite recent. \citewithauthor{LiuLCY15} pack irregular 3D shapes using a Minimum-Total-Potential-Energy (MTPE) heuristic and prioritize shapes with the lowest gravitational center of height. \citewithauthor{WangH19a} propose the Heightmap Minimization (HM) heuristic to minimize the volume increase of packed non-convex shapes as observed from the loading direction. \citewithauthor{GoyalD20} make their packing decision via the Bottom-Left-Back-Fill (BLBF) heuristic~\cite{tiwari2010fast}, which places an object in the bottom-most, left-most, and back-most corner. All these placement rules are manually designed based on specific observations, which limits their applicability. For example, the HM heuristic performs quite well for non-convex shape packing, since it aims to minimize space occupancy vertically. However, it cannot differentiate horizontal placements on a flat container, leading to sub-optimal cases demonstrated in~Fig.~\ref{fig:Insight}a.

The latest efforts~\cite{hu2017solving,DuanHQGZWX19,HuXCG0020,ZhaoS0Y021,zhao2022learning} resort to machine learning, specifically RL, to automatically synthesize policies that work well on an arbitrary \zhf{cubical shape} packing task without human intervention.
Early works~\cite{hu2017solving,DuanHQGZWX19,HuXCG0020}, however, only learn the order of \zhf{cuboids} to be packed, while using separate heuristics or algorithms to compute the placement. More recent work~\cite{ZhaoS0Y021} learns the placement policy via predicting logits of discretized placement locations. The latest work~\cite{zhao2022learning} combines the merit of heuristic and learning-based policies using a candidate selection policy parameterization, having the policy rank a pre-defined set of candidate placements. 
Similar candidate selection approaches have been proven effective in a variety of robot manipulation tasks, including grasping~\cite{mahler2016dex}, object transferring~\cite{zeng2020transporter}, and assembly discovery~\cite{FunkCB021, FunkMC022}, where manually designed heuristics are used to prune sub-optimal actions and learned policies further selects promising ones. Our policy inherits these ideas with necessary modifications toward \zhf{irregular shapes}.
A concurrent work is~\cite{HuangWZL23} which investigates a similar problem of learning-based irregular shape packing. Unlike their method which works with a large resolution-complete action space, we propose a candidate selection algorithm to pre-select a small discrete set of candidate placements. 
This makes our approach achieve faster training and better packing utility.

\subsection{Packing in Computer Graphics} There is a handful of packing-related graphic problems, most of which involve irregular shapes and continuous decision variables. UV atlas generation, for example, minimizes the memory footprint by packing many texture charts into a single texture image. To generate high-quality textures, an algorithm needs to jointly optimize irregular chart shapes and their continuous packing placements. Early works~\cite{levy2002least,10.2312:egs.20031064} nail down the two classes of decision variables in separate sub-stages. More recent works~\cite{LimperVS18,Liu_AAAtlas_2019,ZHANG2020101854} couple the two sub-stages by decomposing the charts for higher packing efficacy. \citewithauthor{LimperVS18} propose a heuristic strategy to iteratively cut charts into smaller ones and repack them using \zhf{heuristics}~\cite{noll2011efficient}. However, the shape of chart boundaries is pre-determined and cannot be modified during cutting. \citewithauthor{Liu_AAAtlas_2019} and \citewithauthor{ZHANG2020101854} propose to deform charts into rectangular patches and then adopt rectangular packing heuristics~\cite{schertler2018generalized}. But their work is irrespective of the sequential nature of the packing problem, deforming chart shapes in a sub-stage. 

A similar requirement to UV atlas generation arises in 3D printing, where the limited volume of commodity 3D printers requires large objects to be partitioned and/or repacked for printing efficacy. The packing plan can be further constrained for manufacturability and structure robustness. The first work~\cite{luo2012chopper} in this direction cuts large objects using the BSP-tree, such that each leaf node fits into the printing volume. They choose cutting planes from the object surface patches, essentially discretizing the continuous solutions. \citewithauthor{luo2012chopper} select BSP-tree structures guided by a score function incorporating various constraints.
However, their optimization strategy is myopic, i.e., using a horizon equal to one. Their follow-up work~\cite{yao2015level} further allows irregular part shapes to be locally optimized for structure robustness and collision resolution, and then packs the parts into the printing volume using local numerical optimization. As a result, \citewithauthor{yao2015level} allow continuous optimization of both irregular object shapes and packing placements, but their optimizer is still myopic. A similar approach has been proposed by~\citet{MaCHW18}, which only optimizes the final object placement result and neglects the packing process.

Our problem is closely related to these works by optimizing online placements for general 3D shapes. Modeling  packing as a sequential-decision making problem, our method bears the potential of closing the sub-optimality gap as illustrated in~Fig.~\ref{fig:Insight}b. A related work to ours is TAP-Net~\cite{HuXCG0020}, which considers sequential object selection and packing problem in a discrete state-action space, assuming cubical objects and perfect placement without physics dynamics. Our method deviates from~\cite{HuXCG0020} in two important ways. First, we assume continuous object placements and physics realizability constraints. This is a much more realistic setting mimicking real-world packing problems. 
Our physics simulator allows uncertainty to be modeled, lifting the assumption of perfect policy execution. Second, our policy parameterization enables the RL algorithm to directly train the entire policy end-to-end, instead of only the object selection policy as done in~\cite{HuXCG0020}, which is \zhf{one} reason for our superior performance over all existing baselines.

\begin{figure*}[ht]
\includegraphics[width=1.0\textwidth]{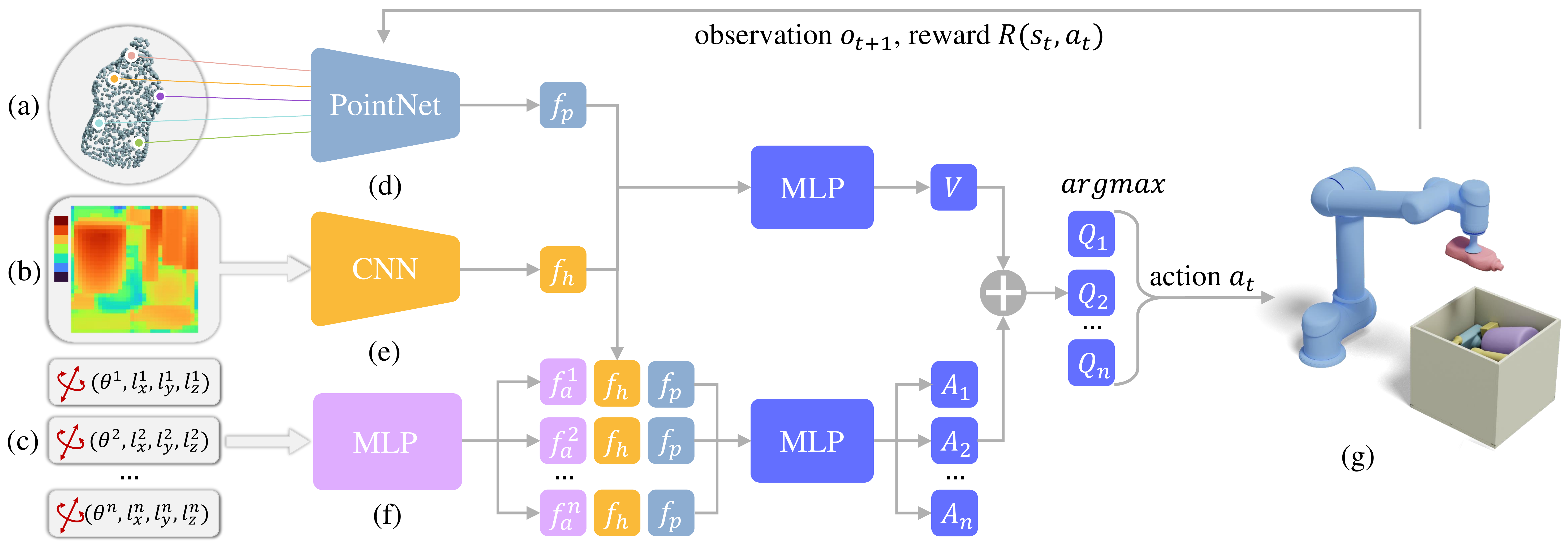}
\caption{\label{fig:architecture} Our policy learning architecture. The input to our method is the surface point cloud (a) of the incoming object, $P$, and the heightmap (b) of continuous object configurations in the target container, $H_c$. Our neural network policy uses PointNet (d) and CNN (e) to extract features of $P$ and $H_c$ respectively for 3D geometry understanding. Our geometric-inspired candidate generalization method would then provide a set of placements (c), each encoded as a feature $f_a^i$ using an MLP (f). Finally, our policy which is a dueling network ranks the placement candidates via the state-action value function $Q$, and the best candidate is selected for execution. The continuous object configurations inside the target container are governed by a physics simulator (g). The packing process continues with receiving the next observation until the container is full. Our RL algorithm trains the ranking policy by asynchronously sampling trajectories and updating policy parameters with  granted reward signals.}
\end{figure*}
\subsection{Packing in Robotics} Roboticists consider packing as a component in the pipeline of perception, planning, predicting, and control, rather than a standalone algorithmic problem. However, the progress in robotic packing tasks is relatively slow. This is because the robust execution of high-quality packing plans is extremely difficult due to the tiny spaces between objects. We are only aware of a few prior works~\cite{WangH19a,8793966,FunkCB021} presenting full-featured packing systems. \citewithauthor{WangH19a} use a robot arm with a sucker-type gripper to grab objects from top-down views. Their following work~\cite{WangH22}  further introduces a fallback strategy to shake the target container and create new open spaces for packing. \citewithauthor{8793966} restart the sensing, motion planning, and control loop whenever failures are detected in the downstream stages. Both methods use simple heuristics to solve the underlying packing problem, with \citewithauthor{WangH19a} relying on the heightmap minimization heuristic and \citewithauthor{8793966} assuming known cubical object shapes and compatible target container sizes. 
\citewithauthor{FunkCB021} proposed a novel system that effectively assembles 3D objects into a predetermined structure. For the assembly task, however, their method only assigns the mechanical parts to pre-defined positions of the target structure, while we need to optimize the object poses to maximize the space utility.

In contrast, bin-picking~\cite{mahler2016dex,mahler2017learning} can be robustly executed on robot hardware, because the target container is assumed to be much larger than objects, rendering packing unimportant. The bin-picking solutions~\cite{mahler2016dex,mahler2017learning}, though quite different from ours, share commonality with our policy design. Both methods assume the availability of a set of candidate actions, which is then ranked by a learning-based policy. Most recently, \citewithauthor{zeng2020transporter} and \citewithauthor{huang2022equivariant} bring the accuracy and generality of learning-based object transfer policy to another level by introducing equivalency. By factoring out the 2D rigid rotations from the neural network input-output space, learning becomes much more sample-efficient. We adopt a similar approach for \zhf{factoring out rigid rotations } during the forward calculation of  packing policies.
\section{\label{sec:method}Method}
We introduce our online packing problem setup in~Section~\ref{sec:definition} and formulate it as Markov Decision Process (MDP) in~Section~\ref{sec:environment}. 
To effectively solve this problem, we design a novel packing pattern based on candidate actions generated by a theoretically-provable method  in~Section~\ref{sec:policy}. In~Section~\ref{sec:RL}, we describe our asynchronous RL algorithm for accelerating packing policy training in the physics simulation world. Extensions to buffered packing scenarios will be discussed in~Section\ref{sec:buffer}. The pipeline of our learning-based algorithm is outlined in~Fig.~\ref{fig:architecture}.

\subsection{\label{sec:definition}Problem Statement}
Irregular shape packing problem considers a finite set of $N$ geometric objects $G_1,\cdots,G_N$. Each $G_i\subset\mathbb{R}^3$ (in its local frame of reference) is of an irregular and possibly non-convex shape. Following the conventional Bin Packing Problem~\cite{MartelloPV00} (BPP), the target container $C\subset\mathbb{R}^3$ takes up the space $[0,S_x]\times[0,S_y]\times[0,S_z]\subset\mathbb{R}^3$. 
The goal is 
to move as many objects into $C$ in a collision-free and physically realizable manner  and maximize the packing utility: 
\begin{align}
\label{eq:object}
\sum_{G_{i} \subset C} |G_{i}|/|C|, 
\end{align}
where $|G_{i}|$ and $|C|$ are the volume of object  $G_{i}$ and the container volume, respectively. 

\begin{figure}[t!]
\centering
\includegraphics[width=0.48\textwidth]{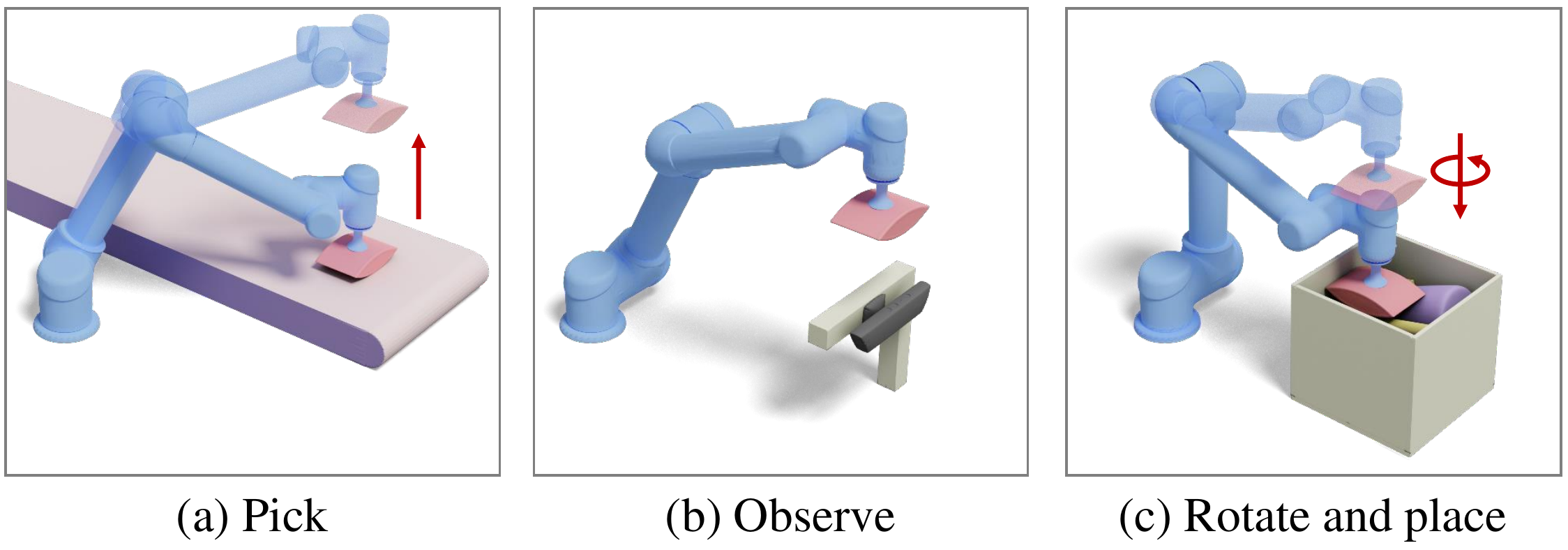}
\vspace{-20pt}
\caption{\label{fig:manipulation} 
Snapshots of robot manipulations. The robot picks up  the incoming object (a) from the conveyor belt and moves this object above an up-looking camera to observe the bottom surface (b). Then, the robot adjusts the vertical orientation of the object and places it in the container (c). The robot only picks and places an object from a top-down view.
}
\end{figure}

\subsubsection{Environmental Setup} To mimic the real-world scenario of pick-and-place tasks, we assume that objects are stably laid on the conveyor belt moving at a constant speed, leaving the robot with a limited time window to pick up and place each $G_i$ into $C$. 
Clearly, the time window size is a complex function of conveyor belt speed, 
robot motion speed, and various system delays. 
However, these factors can be tuned for different hardware platforms and are out of the scope of this work. We assume a fixed time window size, which allows us to model object packing as a sequential-decision problem. 
We postulate that the robot arm is equipped with a sucker-type gripper that can only pick and place an arbitrarily-shaped object from the top, as illustrated in~Fig.~\ref{fig:manipulation},  
and that any object and any placement location in $C$ can be reached by the robot. The robot can also apply a 1D  rotation of the gripper around the Z-axis (vertical), the same assumption is adopted  in~\cite{zhao2022learning}.
Therefore, the space for robot decision is $\mathbb{R}^3\times SO(1)$, consisting of a 3D position and an in-plane rotation. 
As compared with~\cite{HuXCG0020} where the virtual robot can grasp a box from its four sides, our assumptions limit robot mobility but make it more amenable to real-world deployment. 

To make packing decisions, the robot is equipped with three RGB-D cameras to fully observe the packing environment (Fig.~\ref{fig:teaser}a).
The on-conveyor camera captures the top surface of the incoming object while the on-container camera observes the continuous packing configurations inside the container.
After picking up the incoming object, the robot moves it over a third, up-looking camera to capture its bottom surface (Fig.~\ref{fig:manipulation}b), thus fully observing the geometry of the object and applying an additional transformation to pack it.

Different from those works which only optimize the final packing result~\cite{MaCHW18}, PRP also concerns the packing process, i.e., moving the objects into the container one by one.
Since the robot can only observe one object at a time, our packing problem follows the \textit{online} setting~\cite{Seiden02} where each object is placed without the knowledge of the next ones. No further adjustment, such as unloading or readjusting, will be allowed.
The robot must make an immediate decision that accommodates the incoming object while optimizing the overall compactness of the in-container layout. Packing alone is already a difficult combinatorial optimization problem, the arbitrarily complex object geometry and the imperfect placement make this problem even more challenging with huge solution space. We resort to RL to learn this packing skill automatically 
through trial and error.

\begin{figure}[t!]
\centering
\includegraphics[width=0.48\textwidth]{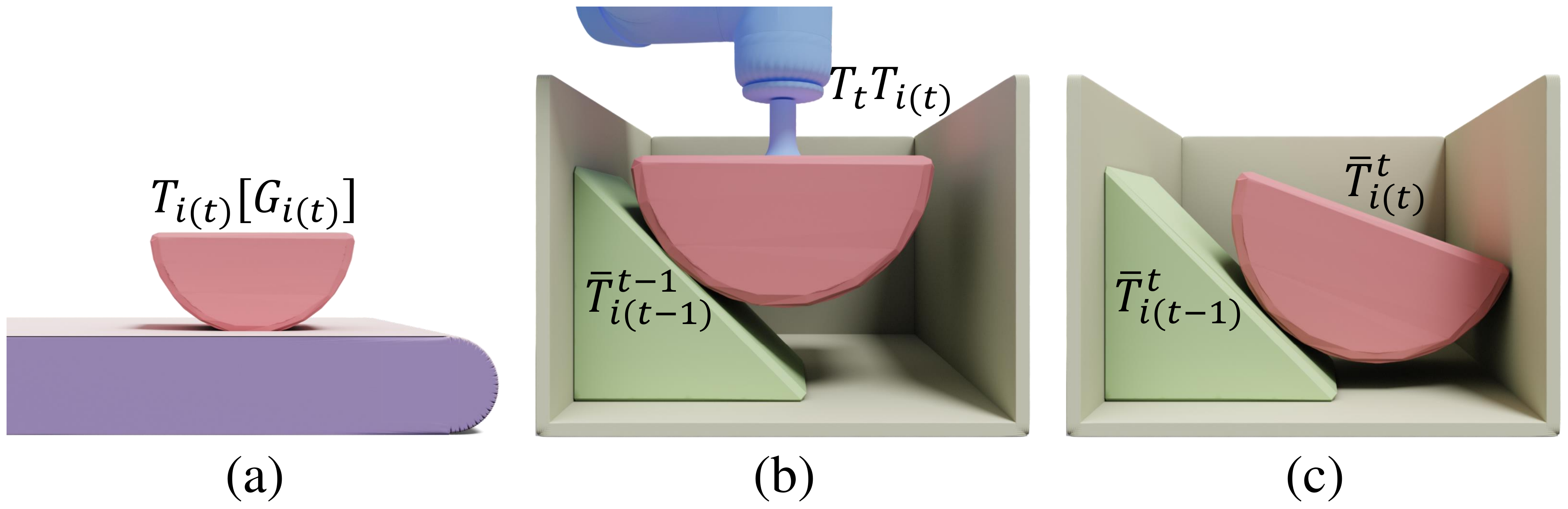}
\vspace{-15pt}
\caption{\label{fig:stable} (a): The object $G_{i(t)}$ is transported by the conveyor with its planar-stable pose $T_{i(t)}[G_{i(t)}]$, at time step $t$. (b): The robot moves this object into the container and releases it with transform $T_tT_{i(t)}$. (c): Governed by rigid body dynamics, the ultimate continuous configuration for $G_{i(t)}$ is $\bar{T}_{i(t)}^t$.}
\vspace{-10pt}
\end{figure}
\subsubsection{Stable Object Poses} As compared with BPP problems which only consider 6 axis-aligned orientations of cuboids, an irregular 3D  object can have an  arbitrary local-to-world orientation spanning the entire $SO(3)$.
However, we can adopt a mild assumption that objects are stably lying on the conveyor belt with force equilibrium. 
This assumption holds when the conveyor belt has sufficient frictional forces and moves reasonably slowly, which is generally the case.
We denote the incoming object id at the $t$th timestep as $i(t)$.
Mathematically, $G_{i(t)}$ is subject to a world transform $T_{i(t)}$ such that $T_{i(t)}[G_{i(t)}]\subset\mathbb{R}^3$ is a physically stable pose on the conveyor belt as~Fig.~\ref{fig:stable}a. The robot can then apply a vertical rigid transformation $T_t\triangleq T(\theta,l_x,l_y,l_z)$ such that $(T_tT_{i(t)})[G_{i(t)}]\subset C$ and $(T_tT_{i(t)})[G_{i(t)}]$ is collision-free with the boundary of $C$ or any other objects already in $C$, as shown in~Fig.~\ref{fig:stable}b. Here $\theta$ is the vertical \zhf{rotation angle} and $(l_x,l_y,l_z)$ \zhf{the} 3D target position which takes the Front-Left-Bottom (FLB) of the Axis-Aligned Bounding Box (AABB) of $G_{i(t)}$ as the reference point. Finally, the robot releases $G_{i(t)}$ with $T_tT_{i(t)}$ and the ultimate continuous configurations of all placed objects $G_{i(1)},\cdots,G_{i(t)}$, denoted as $\bar{T}_{i(1)}^t,\cdots,\bar{T}_{i(t)}^t$, are governed by rigid body physics dynamics like~Fig.~\ref{fig:stable}c. 

\subsection{\label{sec:environment}PRP Learning Environment}
In this section, we discuss our PRP environment compatible with all statements of~Section~\ref{sec:definition}. This involves a random packing problem emission procedure and an environment  model that casts online packing as a Markov Decision Process for RL training.

\paragraph{Problem Emission} 
We aim at learning packing skills for 3D shapes of a specific data distribution, so the training and testing sequences are generated with objects from the same object dataset. The sequences are randomly generated and are not shared by training and testing. Although this is a typical assumption by most policy learning works,
we will demonstrate the generalization of our method to out-of-distribution objects in~Section~\ref{sec:generalization}.
Given a shape set, we emit a random packing problem (a sequence of posed objects) by first picking a random shape from the dataset and then selecting a stable pose for it, both with uniform probability and bootstrap sampling. Then, the vertical in-plane orientation of an object is also uniformly randomized. The sampling is repeated until the objects of the sequence are enough to fill the container ($\sum_i{|G_i|}>|C|$).

\paragraph{Markov Decision Process}
Our online packing problem can be formulated as a Markov Decision Process, which is a tuple $<\mathcal{S},\mathcal{A},\mathcal{P},R,\gamma>$. During each step $t$ of the decision, an agent observes the (partial) state $s_t\in\mathcal{S}$ of the current environment and makes a decision $a_t\in\mathcal{A}$, where $\mathcal{S}$ and $\mathcal{A}$ are the state space and the action space, respectively. The environment then responds by bringing $s_t$ to $s_{t+1}$ via a stochastic transition function $s_{t+1}\sim{\mathcal{P}}(s_t,a_t)$ and granting a reward $R(s_t,a_t)$. The agent is modeled as a policy function $a_t\sim\pi(o(s_t),\omega)$, where $o$ is the observation function and $\omega$ the parameter of $\pi$. Under this setting, solving the online packing problem amounts to the following stochastic optimization:
\begin{align}
\label{eq:MDP}
\text{argmax}_{\omega}\quad\mathbb{E}_{s_0\sim I,\tau \sim \pi,\mathcal{P}}
\left[\sum_{t=0}^{\infty}\gamma^tR(s_t,a_t)\right],
\end{align}	
where $I$ is the stochastic problem emitter,  $\gamma$ is a constant discount factor, and $\tau = (s_0, a_0, s_1, ...)$ is a sampled trajectory.
Below we postulate each component of our MDP, putting together to form our packing environment.

\paragraph{State Space $\mathcal{S}$ and Transition Function $\mathcal{P}$} At timestep $t$, the true state of the current environment involves ultimate object  configurations inside the target container and the state of the  incoming object, i.e.:
\begin{align*}
s_t\triangleq<\bar{T}_{i(1)}^{t-1}[G_{i(1)}],\cdots,\bar{T}_{i(t-1)}^{t-1}[G_{i(t-1)}],T_{i(t)}[G_{i(t)}]>,
\end{align*}
essentially mimicking the online packing setting.  The robot applies a 
temporary, initial transform $T(\theta,l_x,l_y,l_z)T_{i(t)}$ for the $i(t)$th object. After that, our transition function, aka., the rigid body simulator~\cite{coumans2016pybullet}, then integrates the poses of all $t$ objects to reach force equilibrium. We decide that all objects have reached force equilibrium when they have a velocity magnitude smaller than some threshold. At the force equilibrium state, we check for any objects that fall outside $C$ or have a height beyond $C$, in which case we terminate the episode.

\paragraph{Observation Function \textit{o}}
We provide enough
RGB-D cameras for capturing continuous object configurations inside container $C$ and the top/bottom surface of the incoming object $T_{i(t)}[G_{i(t)}]$. We assume that the captured RGB-D images have been segmented into foreground objects and  background. We discard all color details and only retain the depth information. We further extract the heightmap of container $H_c$ with resolution $\lceil{S_x/\Delta_h}\rceil\times\lceil{S_y/\Delta_h}\rceil$, where  $\Delta_h$ is a regular interval. And we get a surface point cloud $P$ belonging to $T_{i(t)}[G_{i(t)}]$. We thus define our \zhf{observation function} as $o(s_t)=(H_c,P)$, which is also illustrated in~Fig.~\ref{fig:architecture}ab.

\paragraph{Action Space $\mathcal{A}$} 
As mentioned in~Section~\ref{sec:definition}, the space for robot decision 
spans the entire~$\mathbb{R}^3\times SO(1)$, involving desired packing position and vertical orientation of the $i(t)$th object.
We can naturally omit the $z$-dimension decision because of the top-down placement manner.
\zhf{Typically, a robotic hardware platform such as~\cite{9560782} is equipped with force sensors and can determine the opportune time for releasing an object once collisions are detected.
}
Therefore, no height measurement is needed for the object.
Similarly, once the bottom object surface and the container heightmap are captured, we can get the object's landing altitude $l_z$ when being placed at $(l_x,l_y)$ coordinates \cite{WangH19a} and we denote $l_z$ as a function $l_z(l_x,l_y)$.
Our packing policy only needs to figure out horizontal positions $l_x,l_y$ for given vertical rotation $\theta$, essentially reducing the action space to $SE(2)\times SO(1)$. 
We define our action as $a_t\triangleq(\theta,l_x,l_y, l_z)$, where $l_z$ is optional. 

However, the $SE(2)\times SO(1)$ space is still unaffordable for the sequential-decision nature of packing. 
For enabling efficient and effective policy learning, we propose to prune this enormous action space  to  limited placement candidates via a \zhf{geometric-inspired method} and use a parameterized policy to further select the best one.
Our motivation and implementation of this  candidate-based packing pattern will be deferred to~Section~\ref{sec:policy}.

\paragraph{Reward Signal \textit{R}}
Since we aim to maximize the packing utility in~Equation~\ref{eq:object},  we directly grant a reward $R(s_t,a_t)=w|G_{i(t)}|$ proportional to the volume of $G_{i(t)}$ once $G_{i(t)}$ is successfully placed inside the container. Here $w$ is a constant weight. 
Otherwise, the reward is zero and the trajectory is terminated.
To avoid premature trajectory termination and get more step-wise profits, the agent should learn to optimize the packing process 
for accommodating more future possible objects.

\subsection{\label{sec:policy} Candidate-Based Packing Pattern}
In this section, we introduce our packing policy representation, including a theoretically-provable candidate action generation method and a learnable policy for further candidate selection.

\subsubsection{Candidate Action Generation} 
An intuitive attempt for \zhf{acting} in the $SE(2)\times SO(1)$ space is discretizing it using a regular interval, as done in~\cite{GoyalD20,HuangWZL23}. However, this leads to a large action space which also grows exponentially with higher resolutions and larger container sizes.
Meanwhile, the clustered actions with close distances also result in meaningless RL exploration.
Generating an effective action subset with a controllable size is necessary for efficient packing policy learning which is also verified in~\cite{ZhaoZXHX22} for cuboid packing. 

\begin{figure}[t]
    \centering
    \includegraphics[width=0.48\textwidth]{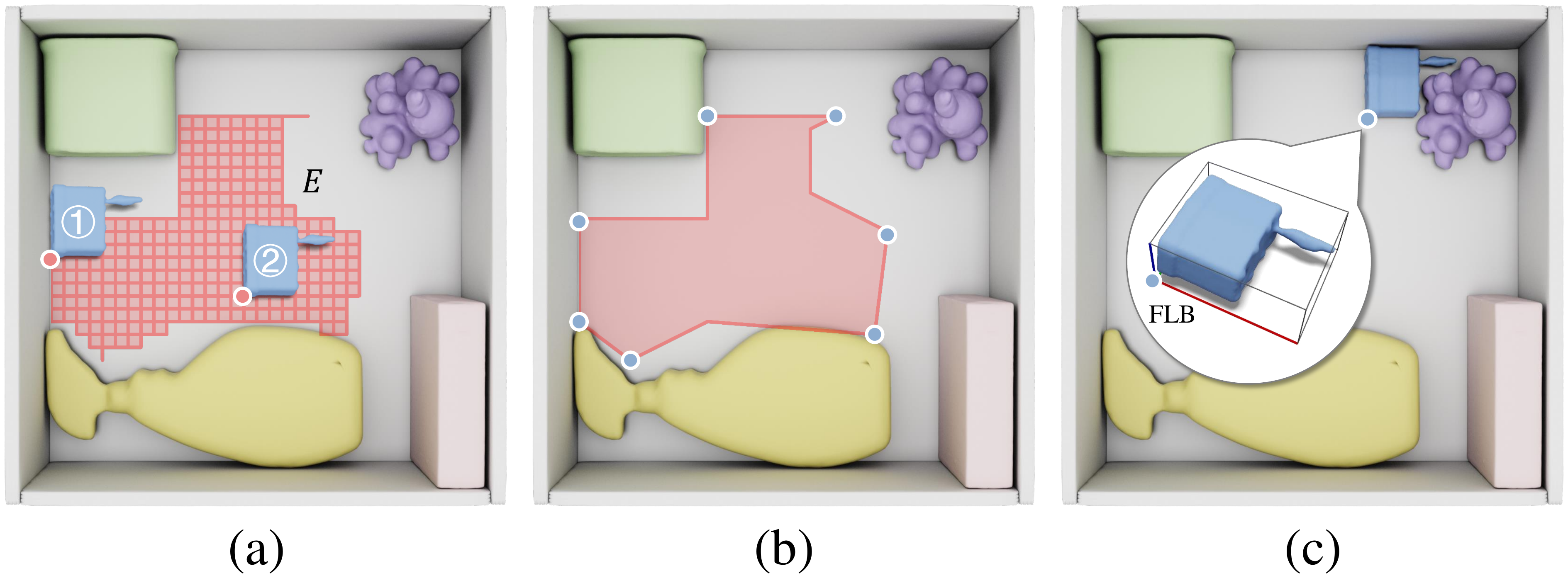}
    \caption{\label{fig:convexAction} We illustrate the procedure for generating candidate actions. Given an in-plane rotation of the object, $\theta$, we first extract feasible and connected action regions $E$ for the incoming object.
    One found region is exemplified in (a). 
    Assigning the incoming object to grid points inside \ding{172} or on the edge \ding{173} of this region would leave small gaps with less potential for accommodating future objects.
    We approximate the contour of this region to a polygon and detect convex polygon vertices (b).
    Our candidate actions correspond to having the FLB corner of the object's AABB at a convex vertex (c) which makes the object placed tightly against the obstacles. This procedure is repeated once for each discretized rotation $\theta$.
    } 
\end{figure}

\begin{figure}[t]
    \centering
    \includegraphics[width=0.48\textwidth]{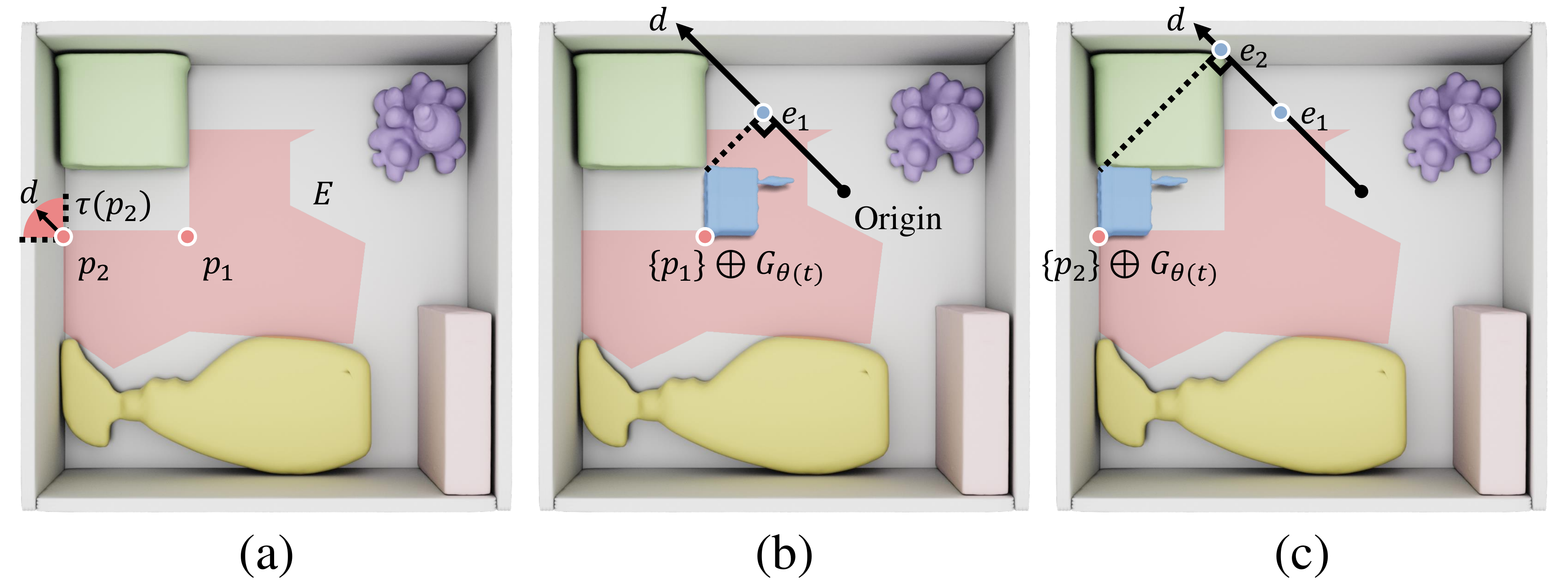}
    \caption{\label{fig:tight_3d} 
    Given a point $p$ in the empty space represented by a polygon $E$ (a), its corresponding placement is $\{p\} \oplus G_{\theta(t)}$ (b).
    All the analysis is performed in 2D by projection and we temporarily replace symbol $G_{\theta(t)}$ with $G$ for brevity.
    We define a tightness measure $\tau(p,G)$ of packing $G$ at $p$ as the range of directions $d$ along which the extreme value of $G$'s projection, i.e., $e(G,p,d)=\max_{g\in\{p\}\oplus G}(d^Tg)$, reaches a local maximal over a neighborhood of $p$: $p=\argmax_{q\in \mathcal{N}(p)}e(G,q,d)$. This means that $G$ is tightly packed against obstacles at $p$ along $d$. In (a), the range for point $p_2$ is depicted as the pink sector.
    For $p_1$ which is a concave vertex, $e_1=e(G,p_1,d)$ is not locally maximal since there are points nearby like $p_2$  which has a larger extreme value $e_2>e_1$. In fact, there does not exist a direction along which $e_1$ can attain local maximal, so $\tau(p_1) = 0$.
    For the convex vertex $p_2$, $\tau(p_2)= \pi / 2$ corresponds to the pink sector in (a).
} 
\end{figure}

Given an object to be packed, we hope to find a finite set of candidate actions that result in the object being tightly placed against the obstacles (container boundary or already placed objects) and permit no further movement, so that maximum  empty room is left unoccupied for future object packing. See a counter-example in~Fig.~\ref{fig:convexAction}a. However, the irregular 3D boundaries of already packed objects can significantly complicate the analysis of candidate actions. To simplify the analysis, we consider the 2.5D top-down height field formed by the packed objects. We first extract near-planar connected regions $\mathcal{E}=\{E_i\}$ formed by areas of similar $l_z$ values. We then consider the boundaries of the projected 2D regions.
To ensure compact packing, we assume that the object should be placed at a (locally) convex vertex of a 2D connected region (Fig.~\ref{fig:convexAction}b), with which we can obtain a discrete set of candidate actions for the object. 
To provide a theoretical validation, we prove the following 2D theorem:
\begin{theorem}
\label{theorem:convex}
For a 2D (polygonal) connected region $E$ and a point on its boundary $p\in\partial E$, if $p$ is a convex vertex, then
$p = \argmax_{q\in \mathcal{N}(p)}\tau(q)$ for an open neighborhood $\mathcal{N}(p)$ of $p$ which does not contain any other convex vertices and a tightness measure $\tau(\cdot)$.
\end{theorem}

We define the tightness measure as the range of directions making $\{p\} \oplus G_{\theta(t)}$ touch the obstacles and permit no further movement, where $\oplus$ is Minkowski sum~\cite{mark2008computational} and we vertically rotate $T_{i(t)}[G_{i(t)}]$ with $\theta$ to get $G_{\theta(t)}$. 
See~Fig.~\ref{fig:tight_3d} for the explanation of this definition. 
We provide a proof of this theorem in the next subsection.
Essentially, this theorem claims that a convex vertex of $E$ is a \emph{local optimum} that leads to a tight object packing (see~Fig.~\ref{fig:convexAction}c) which can be chosen as an action of the  candidate.
We extract candidate actions from all $E_i \in \mathcal{E}$. We employ RL policies to further rank these locally optimal candidate actions and make globally optimal decisions. This strategy effectively reduces the dimension of action space.

\subsubsection{Rationale of Candidate Generation Method}\label{sec:theory}

\begin{figure}[t]
\centering
\centerline{\includegraphics[width=0.48\textwidth]{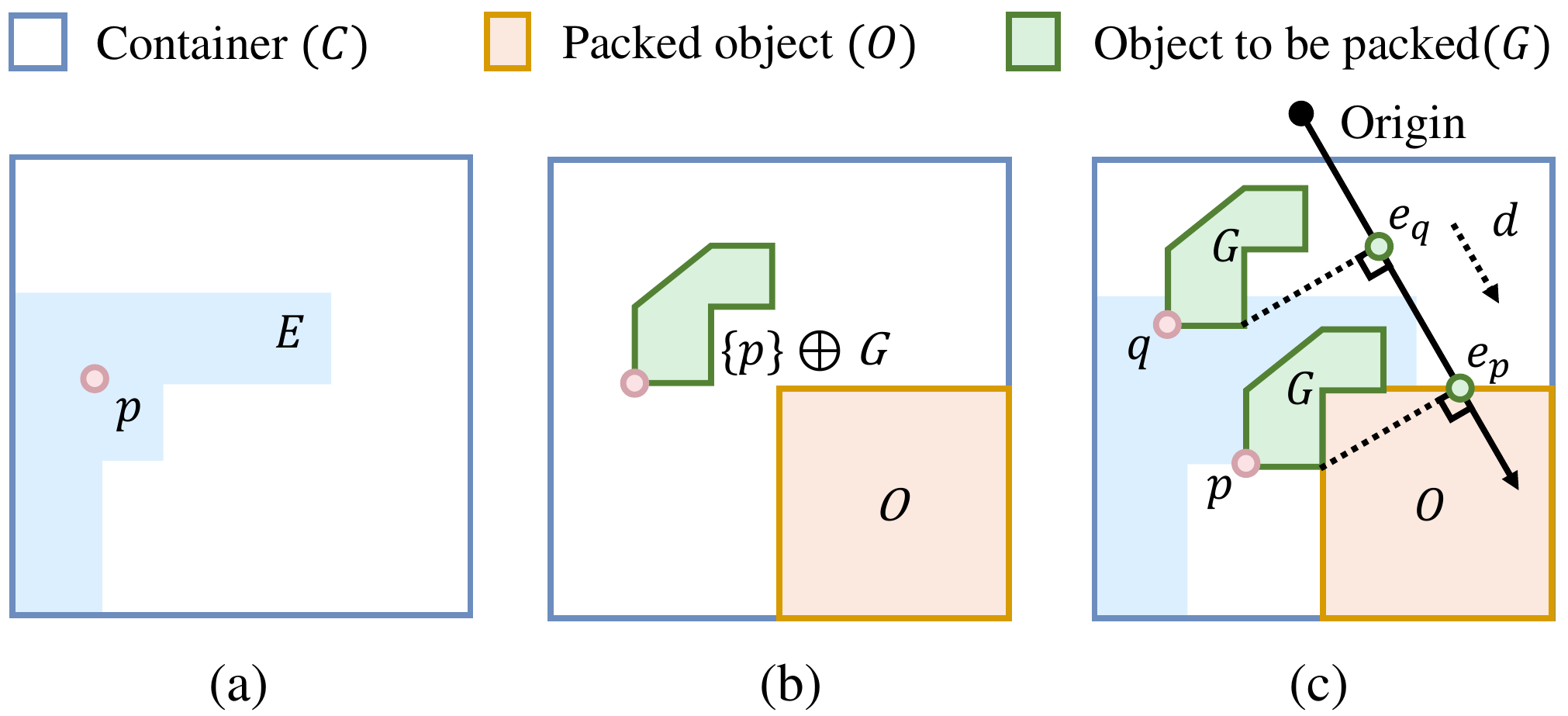}}
\caption{\label{fig:minkov_maintext}
Select a reference point $p$ from the feasible action region $E$ (a), its corresponding placement is $\{p\} \oplus G$ (b). (c):
In direction $d$, the projection $e_p = e(G,p,d)$ of  $\{p\} \oplus G$  is higher than $e_q$ of  $\{q\} \oplus G$.
Thus $\{p\} \oplus G$ is a more extreme placement in $d$. Besides, $\{p\} \oplus G$ is tightly against obstacle $O$ and can no longer move in $d$. Therefore,  $e_p$ is local optima.}
\end{figure}

We will formally define our tightness measure $\tau(\cdot)$ and provide a proof of Theorem~\ref{theorem:convex}. Given a 3D object with known orientation, we can project it on the horizontal plane as a 2D polygon $G$, so all the following analysis are restricted to 2D by projection. For example, we denote $O\subset\mathbb{R}^2$ as the  horizontal projection of packed objects. We can get feasible action regions $E$ which satisfy that $E\oplus G \subset C$ and  $E \oplus G \cap O = \emptyset$, where $\oplus$ is Minkowski sum~\cite{mark2008computational}. With rotation fixed, the placement of $G$ reduces to selecting a reference point $p = (l_x, l_y)$ from $E$ and $\{p\} \oplus G$ is the corresponding placement (see illustrations in~Fig.~\ref{fig:minkov_maintext}ab).

We prefer point $p$ which places $G$ tightly against obstacles, for which there is no  straightforward metric. 
\citet{mark2008computational} define an extreme concept on polygons,
one polygon is more extreme in a direction $d$ than another if its extreme points lie further in that direction.
Inspired by this, we  adopt   $e(G,p,d)=\max_{g\in\{p\}\oplus G}d^Tg$ which is the extreme value of $G$'s projection for the packing task.
Placing $G$ at $p$ is more extreme than one placement $q$ in direction $d$ if $e(G,p,d)$ is higher than $e(G,q,d)$.
When $G$ is tightly packed against obstacles at $p$ along $d$,
 $e(G,p,d)$ also reaches a local maximal, as
illustrated in~Fig.~\ref{fig:minkov_maintext}c.
Therefore, we get an intuitive indicator of tight packing,
i.e.:
\begin{align}
\label{eq:extreme_maintext}
p=\text{argmax}_{q\in \mathcal{N}(p)}\;e(G,q,d)
\end{align}
for some open neighborhood $\mathcal{N}(p)$ of $p$ and some direction $d\in\mathbb{R}^2$.
The following property is an immediate consequence of the Minkowski sum, which establishes a direct connection between packing tightness and reference points in $E$:

\begin{proposition}
    If~Equation~\ref{eq:extreme_maintext} holds for some open neighborhood $\mathcal{N}(p)$, then:
    \begin{align}
       p=\text{argmax}_{q\in \mathcal{N}(p)} d^Tq
        \label{eq:target_maintext} 
        \end{align}
\end{proposition}
In other words, if $d^Tp$ is local optima, i.e. $d^Tp \ge d^Tq$ for all $ q \in \mathcal{N}(p)$, the point $p$ 
corresponds to 
the most extreme placement of $G$ in its neighborhood.
With this tool at hand, we can directly evaluate and compare the potential packing tightness of a point $p$ by looking at the range of directions $d$ that makes $p$ satisfy~Equation~\ref{eq:target_maintext}.  

Such a range of $d$ corresponds exactly to the spanning angle of a 2D normal cone~\cite{boyd2004convex}. The normal cone of a set $C$ at a boundary point $x_0$ is the set of all vectors $y$ such that $y^T (x - x_0) \le 0$ for all $x \in C$. We summarize this property as our definition of tightness measure below:
\begin{proposition}
For a convex polygonal set $E$ and $p\in\partial E$,
 the spanning angle of the normal cone at $p$ is defined as our tightness measure $\tau(p)\triangleq  \pi-\theta$, where $\theta$ is the interior angle of $E$ at $p$. 
\end{proposition}

\begin{figure}[h]
    \begin{center}
    \centerline{\includegraphics[width=0.5
    \textwidth]{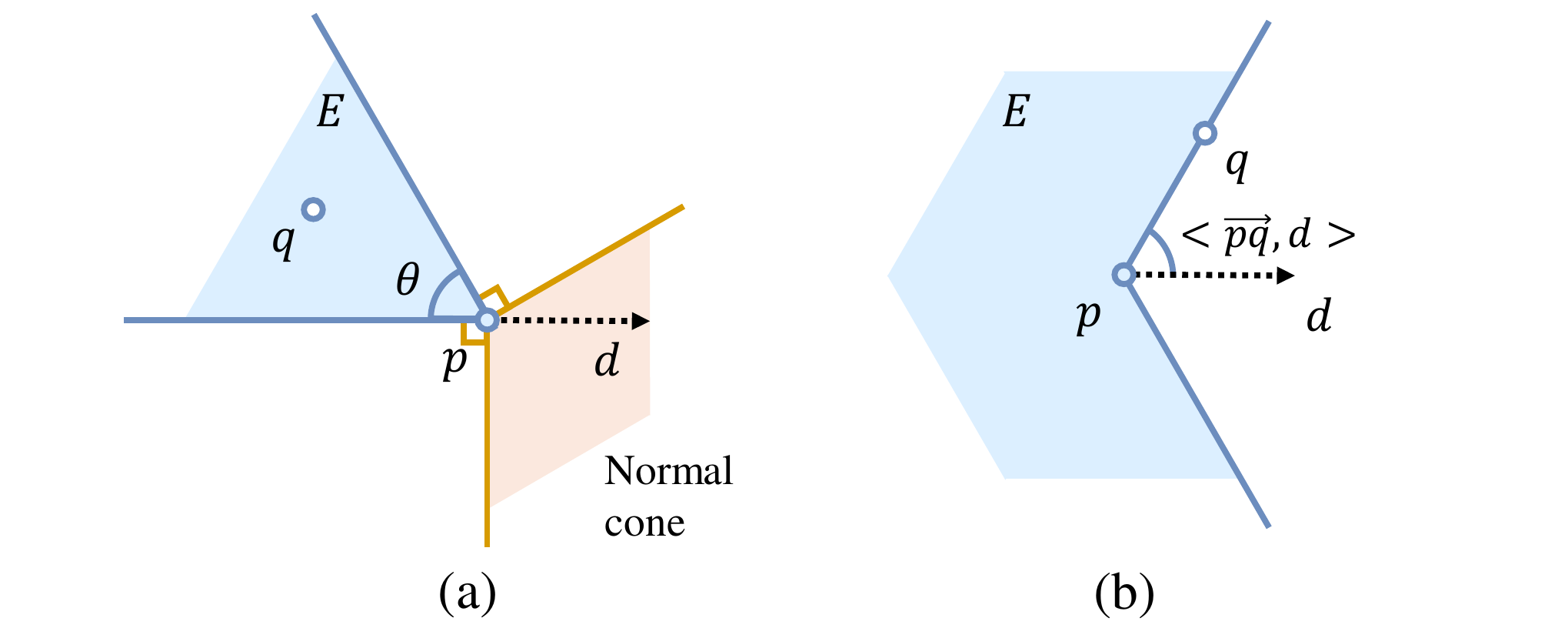}}
    \caption{ 
    (a): For a boundary point $p$ on a convex cone $E$, arbitrary direction $d$ in its normal cone satisfy  $d^Tp \ge d^Tq$, where $q\in E$. (b): If $p$ locates on a concave polygon part, for  $\forall d \in \mathbb{R}^2$, there  exists $q\in E$ that $<\vec{pq},d>$ is less than $\pi/2$, i.e. $d^Tp < d^Tq$. 
    }
\label{fig:proof_maintext}
\end{center}
\end{figure}

A demonstration of a normal cone is provided in~Fig.~\ref{fig:proof_maintext}a. For a concave polygon,
 points located on its concave parts have no normal cone, and we extend that  $\tau(p)=0$ in this case, as shown in~Fig.~\ref{fig:proof_maintext}b.
Using $\tau(p)$ as a tightness metric, we compare different choices of $p$.  
 The proof for Theorem~\ref{theorem:convex} is obvious:
\begin{proof}
Normal cones only exist on boundary points of the polygon and not the  internal ones.
Therefore, we could consider a neighborhood $\mathcal{N}(p)$ containing $p$ and $p$'s two neighboring edges, without any other convex vertices included. Within $\mathcal{N}(p)$, $p$ is convex so $\tau(p)>0$.  All other points $q$ are on a straight line  or concave vertices,  with $\tau(q)=0$. 
\end{proof}
\begin{figure}[h]
    \centering
    \includegraphics[width=0.48\textwidth]{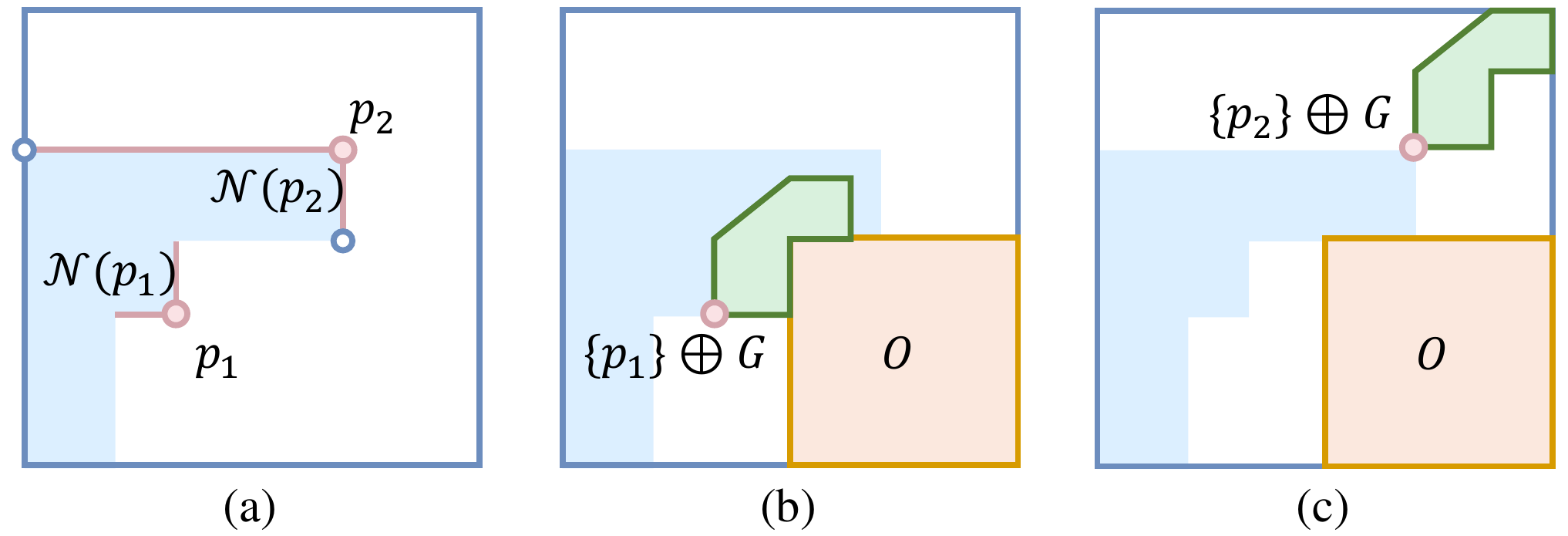}
    \caption{
        (a): The pink parts are neighborhoods $\mathcal{N}(p_1)$ and $\mathcal{N}(p_2)$ for convex vertex $p_1$ and $p_2$. Both $\tau(p_1)$ and $\tau(p_2)$ are local optima and they correspond to object placements, (b) and (c), tightly against obstacles. 
        }
    \label{fig:neighborhood_maintext}
\end{figure}


We demonstrate some  neighborhoods in~Fig.~\ref{fig:neighborhood_maintext}.
Each convex vertex $p$ corresponds to the local optima of  $\tau(p)$ and  a tight object placement against obstacles.
Based on the above observations, our candidate set consists of all convex vertices of the given $E$.

\subsubsection{Implementation and Policy Parameterization}
Given a fixed in-plane rotation $\theta$, we construct the candidate action set for $G_{\theta(t)}$ using three steps. We firstly consider the full placement set $F$ which satisfy $F \oplus G_{\theta(t)} \subset C$ and $F \oplus G_{\theta(t)} \cap \bar{T}_{i(t')}^{t-1} G_{i(t')} = \emptyset$, for $t' < t$. We discretize the container $C$ into regular grids and sample grid points $(l_x,l_y, l_z)$ lying in $F$.
Next, we detect and cluster all feasible, connected 2D regions $\mathcal{E}$ from discretized positions, where each $E \in \mathcal{E}$ results in $G_{\theta(t)}$ placed with similar altitudes, see also~Fig.~\ref{fig:convexAction}a. 
We denote two neighboring grids $(l_x,l_y)$ and $(l_x',l_y')$ as connected if:
\begin{align}
\label{eq:neighbor}
|l_z(l_x,l_y)-l_z(l_x',l_y')|\leq\Delta_z,
\end{align}
with $\Delta_z$ being a constant parameter. As our third step, for each connected region $E \in \mathcal{E}$, we draw the region contour $\partial E$~\cite{SuzukiA85} where $\partial E \oplus G_{\theta(t)}$ touches the container boundary or packed objects from the top-down view.
Since contours $\partial E$ are pixelized, 
we approximate $\partial E$ to polygons with the Ramer-Douglas-Peucker algorithm \cite{Ramer72} and detect convex polygon vertices as  candidate FLBs (Fig.~\ref{fig:convexAction}b).
We execute this procedure for all possible in-plane rotations discretized at a regular interval of $\Delta_\theta$. Finally, the number of candidate FLB positions can be large, and we sort them by $l_z(l_x,l_y)$ in ascending order and retain the first $N$. 
We outline  the candidate generation details in~Algorithm~\ref{alg:candidate}.
Our candidate generation procedure imposes no requirement on object geometry and fits the general shape packing need.
  
\begin{algorithm}[t]
\caption{\label{alg:candidate}Candidate Action Generation}
\begin{algorithmic}[1]
\STATE Sample $\theta$ at a regular interval $\Delta_\theta$
\STATE{Sample  points $(l_x,l_y)$ from $H_c$ per $\Delta_g$ grids}
\FOR{each sampled $\theta$}
\FOR{each grid point $(l_x,l_y)$}
\STATE Compute $l_z(l_x,l_y)$ 
\STATE Rewrite infeasible $l_z(l_x,l_y) = \infty$
\ENDFOR
\FOR{each pair of grid points $(l_x,l_y)$ and $(l_x',l_y')$ of $H_c$}
\STATE Connect neighbors if~Equation~\ref{eq:neighbor} holds
\ENDFOR
\STATE Candidate action set $A\gets\emptyset$
\STATE Detect connected regions
\FOR{each connected region $E \in \mathcal{E}$}
\STATE \zhf{Draw the region contour and approximate it as a polygon }
\STATE \zhf{Detect convex polygon vertices $p\in E$
where $p = \argmax_{q\in \mathcal{N}(p)}\tau(q)$} 
and insert them into $A$
\ENDFOR
\STATE Sort $A$ by $l_z(l_x,l_y)$ and retain the first $N$
\ENDFOR
\end{algorithmic}
\end{algorithm}

Given the packing observation \zhf{tuple} $(H_c,P)$ and the generated candidate actions, our packing policy $\pi(o(s_t),\omega)$ is used to understand 3D task geometry and rank candidates for selection.
Our policy first encodes the container heightmap $H_c$ by a Convolutional Neural Network (CNN) to extract feature $f_h$. Similarly, we use a PointNet architecture~\cite{QiSMG17} to project the point cloud $P$ to feature $f_p$. 
Both feature extractors are designed lightweight and trained from scratch. We apply a vertical transform to the point cloud until its AABB has the smallest volume with FLB at the origin. This essentially follows a similar idea to~\cite{zeng2020transporter} \zhf{which} transforms the point cloud to a canonical pose to improve data efficiency. The $i$th candidate action is brought through an element-wise Multi-Layer Perceptron (MLP) to derive a candidate descriptor $f_a^i$. Our candidate selector then takes the same form as a standard dueling Q-network~\cite{WangSHHLF16}, which is also illustrated in~Fig.~\ref{fig:architecture}. This architecture represents the state-action value function $Q(s,a)$ as:
\begin{align*}
Q(s_t,a_i)=V(s_t) + A(s_t,a_i)
\end{align*}
where $N$ is the number of candidate actions. $V(s_t)$ is called the state value function and $A(s_t,a_i)$ is the advantage function. We concatenate and parameterize the problem features to $V$ and $A$ with two MLPs, with learnable parameters $\alpha$ and $\beta$, respectively: 
\begin{align*}
V(s_t)=&\text{MLP}(f_h,f_p,\alpha),\\
A(s_t,a_i)=&\text{MLP}(f_h,f_p,f_a^i,\beta).
\end{align*}
Our ultimate action is defined as:
\begin{align*}
a_t\triangleq\text{argmax}_{a_i}Q(s_t,a_i).
\end{align*}
The number of candidate actions accepted by our policy is fixed to $N$. If the factual number of candidates is less than $N$, we fulfill the candidate array with dummy actions that are all-zero tuples. The Q-value predictions for these redundant candidates are replaced with $-\infty$ during action selection.

\subsection{\label{sec:RL}Asynchronous Policy Training in Simulation World}
Aside from that the complex irregular geometry and imperfect object placement  enlarge the combinatorial solution space of packing, the heavy cost of exploring this enormous solution space, i.e. physics simulation, also makes it more difficult to learn an effective packing policy.
In this section, we discuss several practical approaches to improve the efficiency of exploration and policy training in the simulation world. We choose a data-efficient off-policy RL algorithm and further use an asynchronous pipeline for accelerating the training, where the simulation-intensive trajectory sampling and policy optimization are performed asynchronously; a partial simulation further lowers the sample cost.

%

Existing reinforcement learning algorithms can be largely divided into on-policy approaches~\cite{SchulmanWDRK17,WuMGLB17} and off-policy methods~\cite{Barth-MaronHBDH18,WangSHHLF16}. The former maximizes the expectation in~Equation~\ref{eq:MDP} by sampling trajectories using the current policy, while the latter maintains an experience memory $D$ that stores trajectories sampled using out-of-sync policies. Off-policy methods minimize the Bellman loss using samples from $D$:
\begin{align*}
\text{argmin}_\omega\quad\mathbb{E}_{(s,a,r,s^\prime)\sim D}\left[(r+\gamma\max_{a^\prime}Q(s^\prime,a^\prime)-Q(s,a))^2\right].
\end{align*}
The ability to reuse samples makes off-policy algorithms more data-efficient, which is critical to our problem. As mentioned in~Section~\ref{sec:policy}, we use dueling networks~\cite{WangSHHLF16} to allow more frequent updates of $V(s)$ and share the learning across multiple candidate actions. We adopt the discrete action space DRL algorithm --- Rainbow~\cite{HesselMHSODHPAS18} to train the dueling networks. Besides the dueling architecture, the Rainbow algorithm also fruitfully combines the well-known DQN method~\cite{MnihKSRVBGRFOPB15} with other five independent improvements like prioritized replay~\cite{SchaulQAS15}, noisy nets~\cite{FortunatoAPMHOG18}, distributional Q-learning~\cite{BellemareDM17}, and so on.

\begin{figure}[t!]
\centering 
\includegraphics[width=0.48\textwidth]{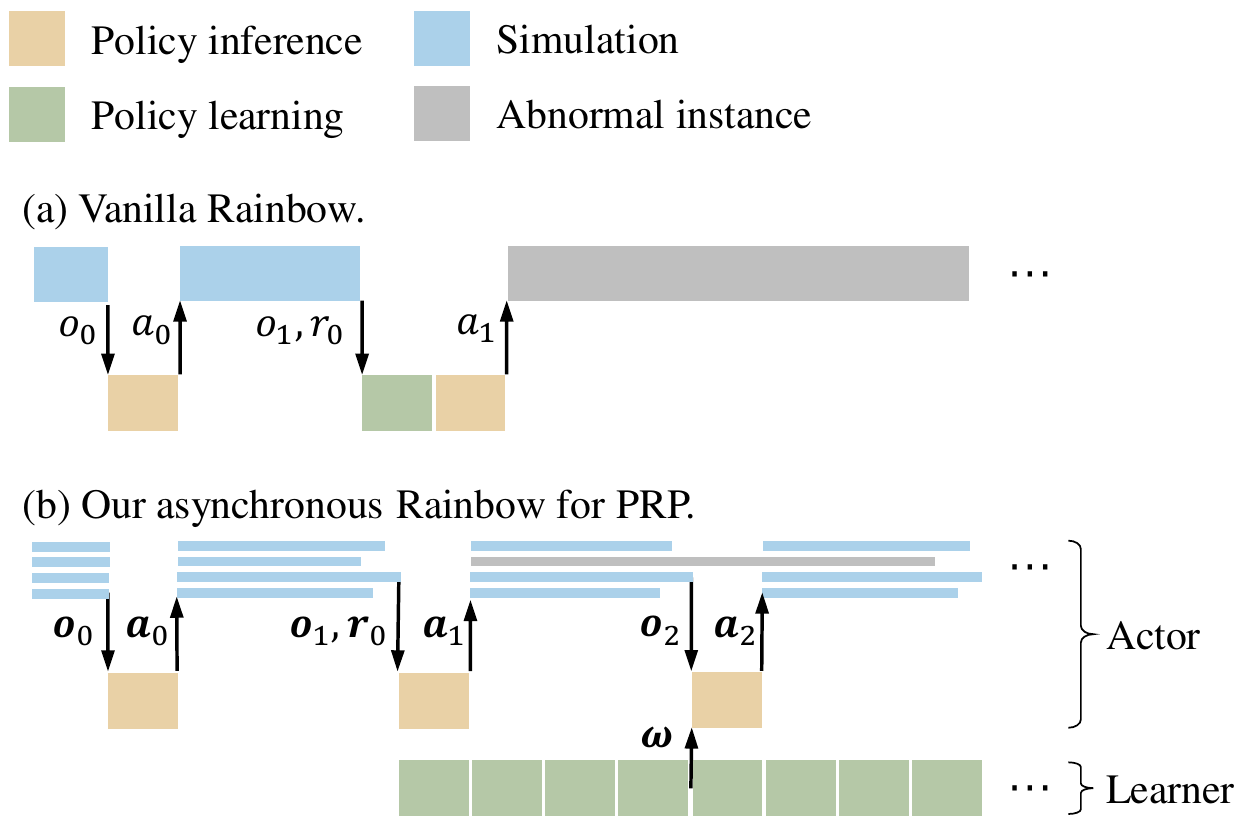}
\caption{\label{fig:DataCollect}
The vanilla Rainbow flowchart (a) and that of our asynchronous version (b). The vanilla Rainbow (a) runs the CPU-bound physics simulations and GPU-bound policy optimization in a sequential manner, which leads to idle computing resources and under-trained policies. This also results in the uniform training schedule being blocked when an abnormal  instance (gray) is sampled.
Instead, our method (b) runs an  experience sampling actor and a policy learner in asynchronous processes. The experience sampling is performed on a batch of CPU threads. The learner keeps learning on GPU and shares the updated policy parameter $\omega$ with the actor. 
The batch of threads is synchronized  when policy inference is needed. We further incorporate an abnormal detection mechanism, so that the abnormal thread is not synchronized with other normal ones. 
}
\end{figure}
\subsubsection{Asynchronous Rainbow} In the vanilla Rainbow algorithm, the experience collection step and the policy learning step run alternatively, as demonstrated in~Fig.~\ref{fig:DataCollect}a. For our problem, however, experience collection via physics simulation is CPU-bound and time-consuming, which leaves policies less trained.
Given that the policy learning is merely GPU-bound, 
we propose to parallelize training via an asynchronous scheme similar to~\cite{HorganQBBHHS18}. Specifically, we create an actor and a learner residing in two different processes, as illustrated in~Fig.~\ref{fig:DataCollect}b. The actor interacts with the packing environment and collects experience. The collected data is saved to a memory $D$ which is shared between processes. The learner keeps learning from $D$ and spontaneously updates the actor with the latest parameters. Our 
asynchronous
implementation not only saves wall-clock training time but also accesses higher-quality experiences by having the actor refine its parameter more frequently before each decision.
 
To further accelerate the experience collection step, we run  multiple simulation threads along with a non-blocking treatment in the actor process. There are two-fold benefits of doing so. First, more sampling threads enrich experiences for policy learning. Moreover, the batched simulation also avoids the uniform training schedule being blocked by abnormally slow instances, as exemplified in~Fig.~\ref{fig:DataCollect}a. Such abnormality can happen when the physical simulator is unstable and experiences a sudden gain in kinetic energy, requiring many more timesteps to converge to a new equilibrium configuration. We suspend the actor thread whenever its reaction time is longer than a threshold, as demonstrated in~Fig.~\ref{fig:DataCollect}b. This suspended thread will then be treated as a standalone instance and rejoin others after its task is finished. This implementation guarantees sufficient concurrency in trajectory sampling.

\subsubsection{Partial Simulation} 
Even by using asynchronous Rainbow, the policy training for PRP is still bound by the massive simulation cost. We can further accelerate training by fixing the configurations of already packed objects. 
In practice, we find that old objects are not affected much by new arrivals. Using this strategy can save a large portion of simulation costs. It can also save the price of computing heightmap $H_c$, since we only need to update the slice of $H_c$ covered by the newly arrived object $\bar{T}_{i(t)}^t[G_{i(t)}]$ but scan the entire container. This simple strategy can significantly boost the simulation frequency by more than three times. Note that this partial simulation is only used during training, and we always simulate all objects and scan the entire container at test time for practicality.

\subsection{\label{sec:buffer}Extension to Buffered Packing Scenario}
\begin{figure}[t]
\centering 
\includegraphics[width=0.48\textwidth]{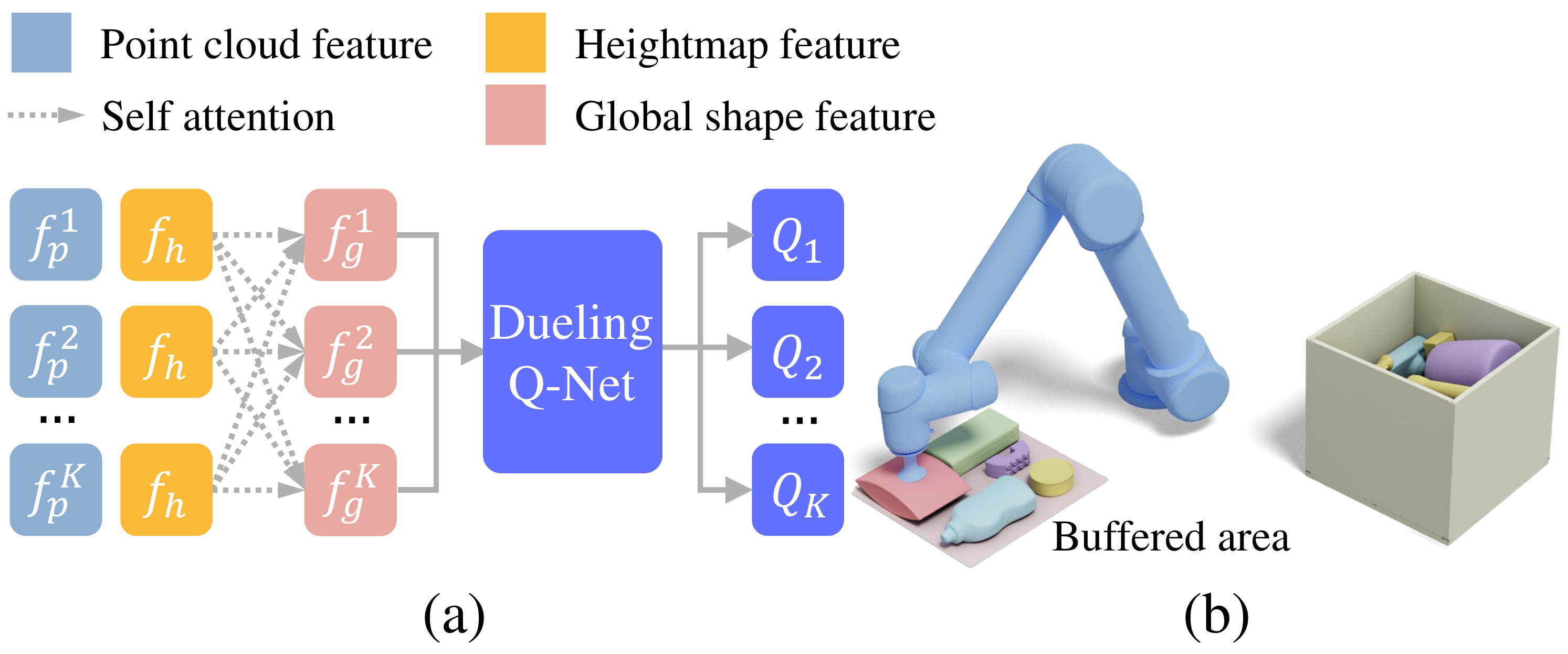}
\vspace{-10pt}
\caption{\label{fig:selector}(a): Our  object-ordering policy, consisting of a Graph Attention Network (GAT) and 
a dueling Q-Network ranker. (b): The robot picks the object with the highest Q-value from the buffer and packs it into the container.}
\vspace{-10pt}
\end{figure}

So far we have discussed the strictly online packing problem. In some real-world scenarios, a buffered area $B$ can be used to temporarily store objects before the final placement, as shown in~Fig.~\ref{fig:selector}.
By introducing the buffer, the robot can reorder objects locally before packing, potentially improving the packing performance by enabling a much larger search space. We suppose the buffered area is of a fixed size $K$. When a new object 
comes, the robot will store this object in $B$. If there are already $K$ objects inside $B$, the robot will pass on one of the objects from $B$ to $C$. Solving this problem not only involves reasoning about  the picked object geometry, but considering the permutation of objects inside the buffer and future possible ones as well. To this end, we use an additional policy $\pi_s$ to select objects from $B$, which is followed by our packing policy $\pi$ in~Section~\ref{sec:environment}. The area of $B$ is sufficiently large such that objects can be placed horizontally, and the point cloud feature of each object in $B$ is stored and available to $\pi_s$. Finally, we assume the robot can reach any position in the buffered area from a top-down view.

Our architecture of $\pi_s$ is illustrated in~Fig.~\ref{fig:selector}a. We assume there are $K$ objects in the buffer, denoted as $G_1,\cdots,G_K$, with their point clouds being $P_1,\cdots,P_K$. Similar to~Fig.~\ref{fig:architecture}, our policy $\pi_s$ first maps each $P_k$ to a feature $f_p^k$ in an element-wise manner and maps the container heightmap $H_c$ to a feature $f_h$. We adopt a Graph Attention Network (GAT)~\cite{VelickovicCCRLB18} to project the tuple $(f_p^k, f_h)$ to a high-level element feature $f_g^k$.
A dueling Q-network block is then used to select the best object with $\text{argmax}_{k=1,\cdots,K}Q(s_t,G_k)$, where $s_t$ is the average of all $f_g$ features.
We train the object-ordering policy $\pi_s$ and the placement policy $\pi$ jointly in an end-to-end manner. That is, $\pi_s$ first chooses one shape $G_k$, then it is $\pi$'s turn to cooperate and choose a  candidate action for placing $G_k$.

\section{\label{sec:result}Results and Evaluation}
In this section, we first explain our packing experiment setup. 
Then, we describe our carefully prepared datasets emulating realistic packing tasks in~Section~\ref{sec:dataPreparation}. 
We illustrate the superiority of our method for online packing by comparing it with a row of existing baselines in~Section~\ref{sec:comparisons} and
demonstrate the benefits of our candidate-based packing pattern in~Section~\ref{sec:actionDesign}.
We report the generalization results of our method in~Section~\ref{sec:generalization}, where the problem emitter transfers to noisy point cloud inputs and unseen shapes. Immediately following, we show performance on buffered packing scenarios in~Section~\ref{sec:buffeResult}.  

We establish our packing environment using the Bullet simulator~\cite{coumans2016pybullet}, with a deep container of size $S_x=32$cm, $S_y=32$cm, and $S_z=30$cm following~\citet{WangH19a}. We assume all objects have the same uniform density for estimating the center of mass. The coefficients of friction among objects and against the container are both set to $0.7$. We set the grid size of the heightmap $H_c$ to be $\Delta_h=1$cm. We then sample $H_c$ at an interval of $\Delta_g=2$ grids to form grid points for the candidate generation procedure. For clustering the connected region with similar $l_z$ values, we use $\Delta_z=1$cm. Unless otherwise stated, we use $\Delta_\theta=\pi/4$ to discretize object rotations. We use a maximum of $N=500$ candidate actions for policy ranking. Our PointNet takes a fixed-sized set of $1024$ points, so we resample $1024$ points from the surface point cloud $P$ to fit this size. 
We adopt $16$ simulation threads in the actor process 
and set the discount factor $\gamma = 0.99$ for training. The training process is conducted on a desktop computer equipped with a Xeon E5-2620 CPU and a TITAN Xp GPU.  Our learning-based policies are optimized within 48 hours with a maximum of 9.6 million interactions with the packing environment. Further analysis of the effects of these experimental parameters and training methods is provided in Appendix~\ref{sec:appendixB}.

\begin{figure}[t!]
\centering
\includegraphics[width=0.48\textwidth]{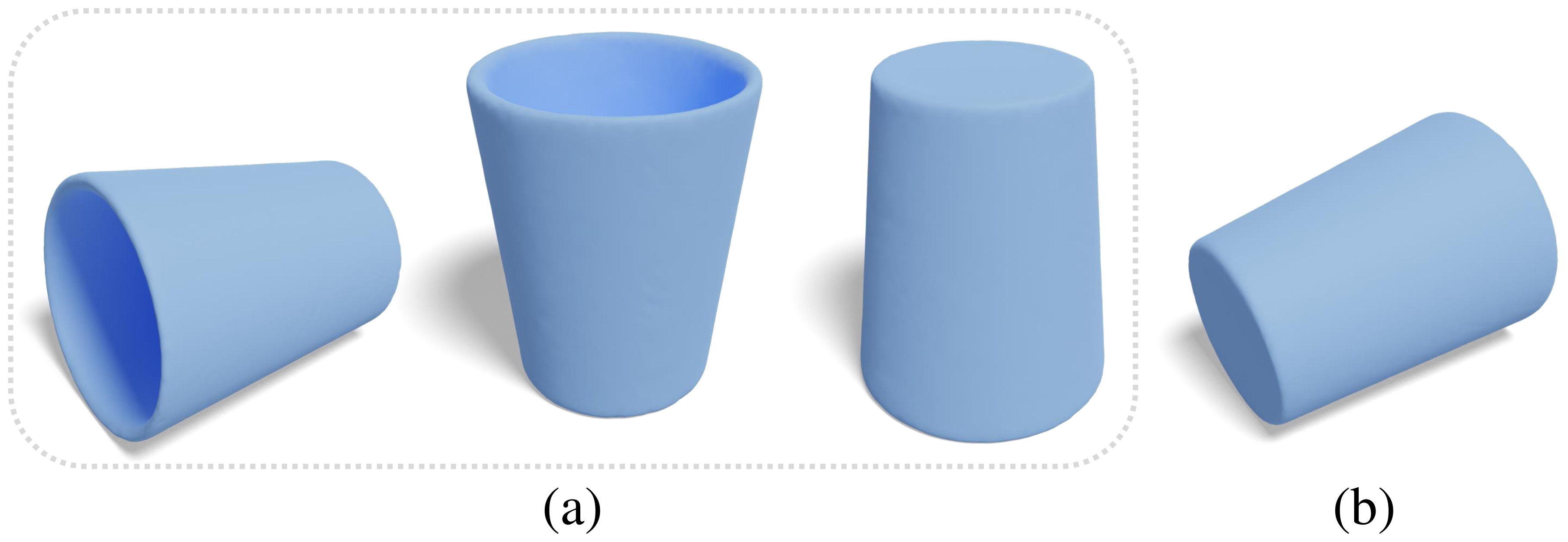}
\caption{\label{fig:poses} (a): Different planar-stable poses of a bucket shape. These poses have invariant appearances after vertical rotations and we remove redundant poses (b) to avoid imbalanced shape distribution.}
\vspace{-16pt}
\end{figure}
\subsection{\label{sec:dataPreparation}Object Data Preparation}
For training reliable policies for packing irregular 3D shapes, we need to prepare object datasets that contain abundant shapes as well as their planar-stable poses, while being compatible with the simulator. 
Our object test suite combines objects from widely used, open-source polygonal mesh~\cite{0033116} datasets, collected either by synthetic modeling or real-world scanning.

To perform robust collision detections and ray-casting tests in the simulation world, we stipulate that objects are represented as closed manifold mesh surfaces. For each object, we first use the method in~\cite{StutzG20} to reconstruct the mesh into a watertight one. We then extract all the planar-stable poses using the detection algorithm~\cite{GoldbergMZCCC99}, as explained in~Fig.~\ref{fig:poses}a. Some planar-stable poses are rotation-symmetric as shown in~Fig.~\ref{fig:poses}b and we propose~Algorithm~\ref{alg:poses} in Appendix~\ref{sec:appendixA} to remove redundant poses and retain only one representative from each rotation-symmetric group. This step avoids unbalanced shape distributions and improves the robustness of trained policies. 

\begin{wrapfigure}{r}{0.48\linewidth}
    \centering
    \vspace{-6pt}
    \includegraphics[width=1.\linewidth]{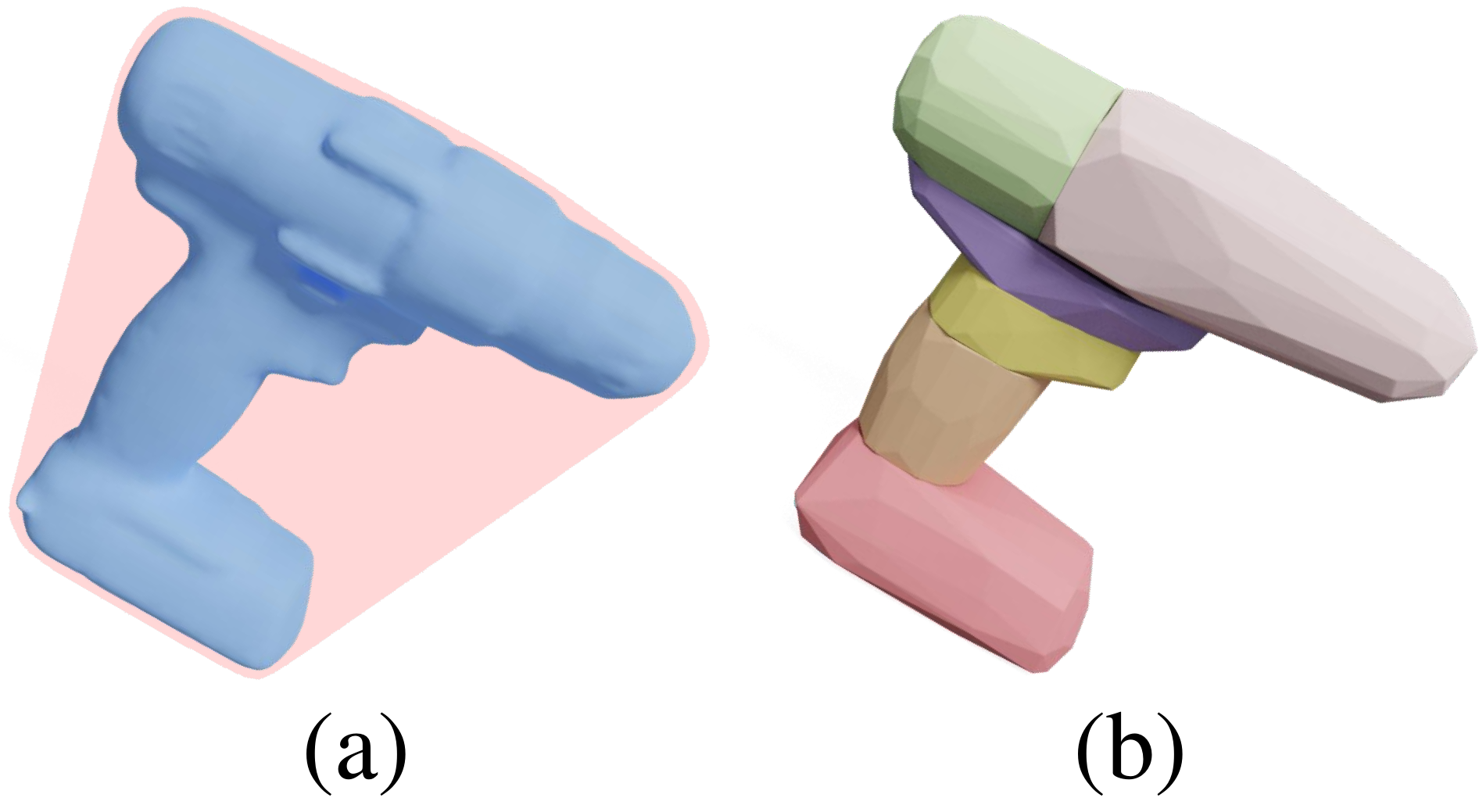}
    \vspace{-20pt}
    \caption{
        Convex decomposition. 
        }
    \vspace{-10pt}
    \label{fig:vhacd}
\end{wrapfigure}  
Finally, several rigid body simulators only accept convex shape primitives, e.g., approximating a non-convex power drill shape with its convex hull in~Fig.~\ref{fig:vhacd}a (red). We thus apply the convex decomposition algorithm~\cite{mamou2016volumetric} for non-convex shapes, as illustrated in~Fig.~\ref{fig:vhacd}b. The complete data preparation pipeline is outlined in~Algorithm~\ref{alg:data} of Appendix~\ref{sec:appendixA}. 
The packing result or the final packing utility is highly related to the geometric 
properties of the picked object dataset. 
For providing convincing evaluations, we apply our pipeline to construct three datasets of \zhf{various} features.


\begin{figure*}[t]
\centering
\includegraphics[width=1.0\textwidth]{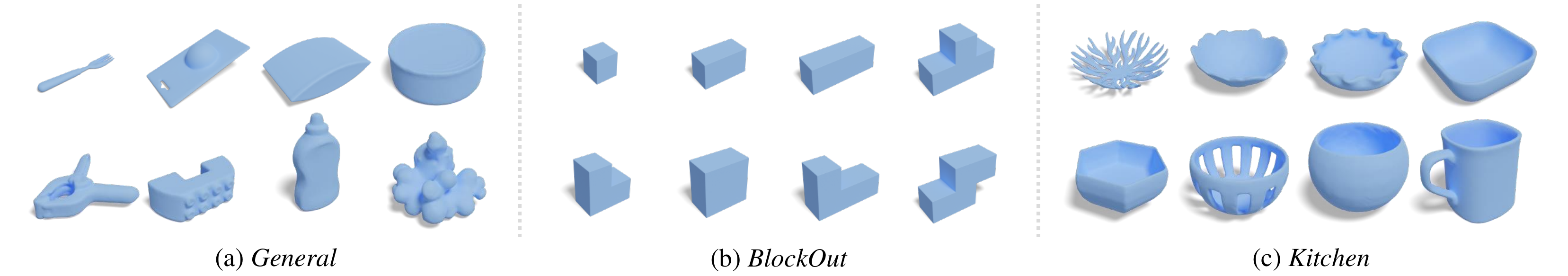}
\caption{\label{fig:dataset} Gallery of our datasets. (a): Part of shapes  from the \textit{General} dataset. (b): All polyominos from the \textit{BlockOut} dataset, where each polyomino is presented with a selected pose. (c): The bowl shapes from the \textit{Kitchen} dataset. The shapes in (a) and (c) are scaled with their maximal AABB size equal to 1.}
\end{figure*}
    
\textbf{\textit{General}} is a large dataset made up of daily life objects, as illustrated in~Fig.~\ref{fig:dataset}a. It combines shapes from the KIT dataset~\cite{KasperXD12}, the BigBIRD dataset~\cite{SinghSNAA14}, the GD dataset~\cite{KapplerBS15}, the APC dataset~\cite{RennieSBS16}, and the YCB dataset~\cite{CalliSBWKSAD17}, totalizing 483 shapes and 1094 planar-stable poses. All these shapes are collected through real-world scanning. \textit{General} is our main dataset for profiling the performance of different methods.


\textbf{\textit{BlockOut}} is a synthetic 3D puzzle dataset with polyominos that comes from~\citet{LoFL09}. This dataset includes 8 polyominos with 23 planar-stable poses, as illustrated in~Fig.~\ref{fig:dataset}b. Each polyomino is composed of basic cubes of the same size. This dataset involves relatively regular shapes that are more complex to handle than cubical objects in BPP problems. 
\textit{BlockOut} 
demonstrates the algorithm's ability to understand shapes and combine them, as done in~Fig.~\ref{fig:Insight}.


\textbf{\textit{Kitchen}} consists of three categories of shapes, namely bowl, board, and fruit. The bowl shapes are concave CAD models collected from the ShapeNet dataset~\cite{ChangFGHHLSSSSX15}, as illustrated in~Fig.~\ref{fig:dataset}c. The board shapes are manually created by us and the fruit shapes are small items coming from the \textit{General} dataset. This dataset is created in order to verify a commonsense logic: an effective packing method  would try to place fruits in bowls before covering  boards. We expect the learned policies to pick up such logical rules without human intervention thus achieving better packing performance.

We compose the polyominos in the \textit{BlockOut} dataset with basic cubes with a side length of $6$cm. \zhf{Therefore, the upper packing utility bound for the chosen container is $87.9\%$.} 
The rotation interval $\Delta_\theta$ for \textit{BlockOut}
is set to $\Delta_\theta=\pi/2$ since the polyomino shapes are axis-aligned.
To construct the \textit{Kitchen} dataset, we collect 54 concave bowl shapes from the ShapeNet
dataset, 106 fruits from the \textit{General} dataset, and 20 planar boards generated with random sizes. The problem emitter for the \textit{Kitchen} dataset is slightly different from the others. 
Since we focus on verifying the packing logic between shape categories, we first sample an object category and then sample a random object with its most stable pose from this category. 
We also provide experimental results on industrial shapes collected from the ABC dataset~\cite{KochMJWABAZP19} in Appendix~\ref{sec:appendixB}.

\begin{figure*}[ht]
    \centering
    \includegraphics[width=1.0\textwidth]{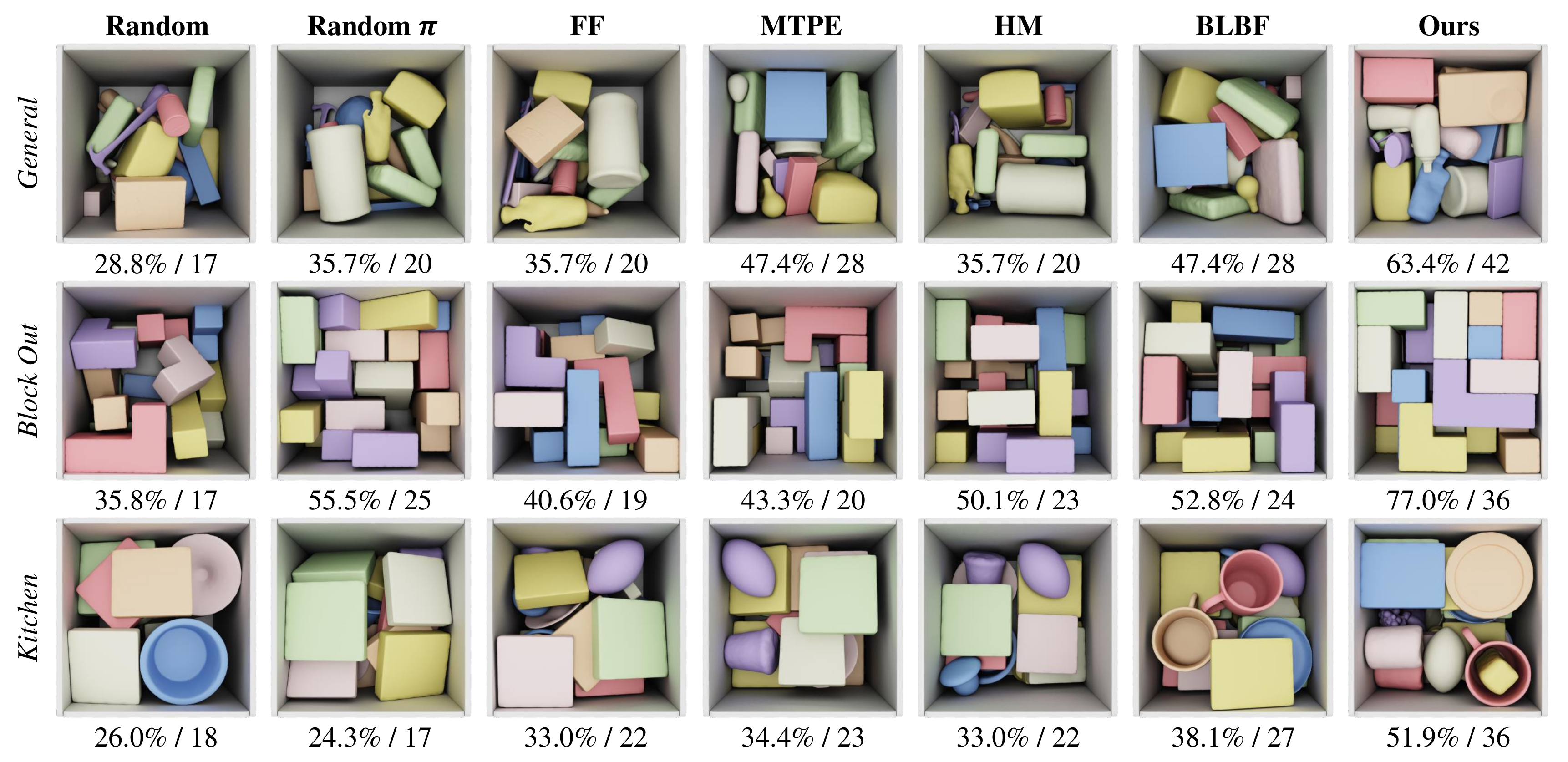}
\caption{\label{fig:visual}Visualization of various packing methods on three datasets. Each test instance is labeled with its utility/number of packed objects. 
    \zhf{Our learned policy consistently exhibits tight packing and achieves the best performance. }
    }
\end{figure*}

\subsection{\label{sec:comparisons}Comparisons with Heuristic Baselines}
We compare our learned policies with several representative  heuristic methods that can pack irregular 3D shapes. 
The MTPE heuristic~\citep{LiuLCY15} searches for the pose with the lowest center-of-mass height. The HM heuristic~\citep{WangH19a} endeavors to minimize the volume increase as observed from the top-down direction. The BLBF~\citep{GoyalD20} heuristic selects  the bottom-most, left-most, and back-most placement. We also test a classic First-Fit (FF)~\citep{Falkenauer96} packing logic which places items in the first found feasible placement.

We run all methods in the same environment setup with the same test sequences. A total of $n = 2000$ object sequences, randomly generated from each dataset, are tested.
The performance of these baselines is assessed across five distinct metrics.
We designate the packing utility as $ u \triangleq \sum_{G_{i} \subset C} |G_{i}|/|C|$ for each test sequence,  which is the volume of packed objects divided by the volume of container $C$. We report the average utility $\bar{u} = \sum_{i=1}^n u_i / n$ to reflect the overall performance of each method.
With the exception of the \textit{BlockOut} dataset where all shapes are composed of regular cubes, optimal packing solutions for other datasets are hard to derive. Consequently, we provide the utility gap against the best baseline. In particular, we calculate $(\bar{u}^* - \bar{u}) / \bar{u}^*$  to highlight the performance discrepancy. Here $\bar{u}^*$ denotes the average utility of the best performer, which is consistently our method. 
We also list the utility variance $\sum_{i=1}^n (u_i - \bar{u})^2 / n$  and the average number of packed objects. Lastly, we present the average computational time for solving each packing problem.
\begin{table*}[t]
\begin{center}
\caption{Online packing performance along five diverse metrics. From left to right: packing utility (Uti.), optimality gap (Gap), the variance of packing utility in units of $10^{-3}$ (Var.), average number (Num.) of objects packed, and average decision-making time measured in seconds (Time).
In addition, we report the statistics of two basic baselines for comparison purposes. The first baseline involves the agent randomly selecting placements from the resolution-complete action space (Random), while the second baseline selects candidate actions using a randomly initialized policy (Random  $\pi$ ) without any specified preference.
Note that, besides the time to find candidates, the decision-making cost of our method also includes  the time required for network inference, which is $4 \times 10^{-3}$ seconds and is consistent across different datasets. 
}
\label{table:baselines}
\setlength{\tabcolsep}{0.6em}
\begin{tabular}{l|crccc|crccc|crccc}
 &  \multicolumn{5}{c|}{\textit{General}} & \multicolumn{5}{c|}{\textit{BlockOut}} & \multicolumn{5}{c}{\textit{Kitchen}} \\
Method & Uti. & Gap  & Var. & Num. & Time & Uti. & Gap  & Var. & Num. & Time  & Uti. & Gap  & Var. & Num. & Time  \\ 
\noalign{\smallskip}
\midrule
\midrule
\noalign{\smallskip}
Random & 
31.6\% & 29.0\% & 5.8 & 19.6 & 0.03 & 
37.1\% & 47.7\% & 3.4 & 17.8 & 0.02 & 
22.6\% & 42.3\% & 6.4 & 16.7 & 0.03 \\
Random $\pi$ & 32.9\% & 26.1\% & 4.6 & 20.4 & 0.04 & 
42.1\% & 40.7\% & 3.5 & 20.2 & 0.02 & 
25.8\% & 34.2\% & 5.5 & 18.9 &  0.03 \\
\midrule
FF~\shortcite{Falkenauer96} & 36.2\% & 18.7\% & 4.6 & 22.2 & 0.03 & 
43.0\% & 39.4\% & 4.0 & 20.6 & 0.02  & 
26.4\% & 32.7\% & 6.3 & 19.3 & 0.03 \\
MTPE~\shortcite{LiuLCY15} & 37.3\% & 16.2\% & 4.6 & 22.7 & 0.04 & 
58.4\% & 17.7\% & 3.6 & 27.9 & 0.02 & 
32.2\% & 17.9\% & 5.1 & 23.4 & 0.03 \\
HM~\shortcite{WangH19a} & 35.8\% & 19.6\% & 4.8 & 21.8 & 0.03 & 
59.5\% & 16.2\% & 3.7 & 28.4 & 0.02 & 
32.6\% & 16.8\% & 5.4 & 23.7 & 0.03 \\
BLBF~\shortcite{GoyalD20} & 36.6\% & 17.8\% & 4.6 & 22.7 & \textbf{0.03} & 
61.9\% & 12.8\% & 4.1 & 29.5 & 0.02 & 
32.0\% & 18.4\% & 5.2 & 23.3 & 0.03 \\
Ours & \textbf{44.5\%} & \textbf{0.0\%} & \textbf{3.4} & \textbf{27.7} & 0.04 & 
\textbf{71.0\%} & \textbf{0.0\%} & \textbf{1.7} & \textbf{34.8} & \textbf{0.02} & 
\textbf{39.2\%} & \textbf{0.0\%} & \textbf{4.5} & \textbf{29.4} &  \textbf{0.03} \\
\end{tabular}
\end{center}
\end{table*}

The quantitative and qualitative comparisons are summarized in~Table~\ref{table:baselines} and~Fig.~\ref{fig:visual}, respectively. The best-performing heuristic for each dataset differs due to their distinct geometric properties. For instance, MTPE has an advantage on the \textit{General} dataset, while HM is the best heuristic for the \textit{Kitchen} dataset which has relatively more non-convex shapes. Compared to these manually designed packing rules, our learned policies adapt well to each dataset and consistently dominate heuristic rules.
The gap metric demonstrates that our method outperforms even the best-performing baseline on each dataset with at least 12.8\% in terms of  packing utility.
All these methods meet real-time packing requirements with a framerate of at least $25$. Moreover, our method achieves the least variance on each dataset, further attesting to its robustness. More packing visualizations are provided in Appendix~\ref{sec:appendixB} for a more detailed understanding. We have also incorporated a dynamic packing video in the supplemental material for a comprehensive perspective.

During the packing process, the number of candidate actions can be large. The cardinality of the candidate set is outlined  in~Table~\ref{table:candidatenum}. In order to mitigate the computational burden, we feed at most $500$ candidate actions to the policy network for each decision. As demonstrated in~Table~\ref{table:baselines}, we can see that randomly selecting a candidate action (Random $\pi$) outperforms picking a random resolution-complete action, while applying our learned policies to rank these locally optimal candidate actions enhances the packing performance even further. 
We have also compared with the baseline that randomly selecting an action from all the candidates  (instead of limiting to at most $500$). The outcomes are  comparable, yielding a performance of 32.9\% for \textit{General}, 42.2\% for \textit{BlockOut}, and  25.5\% for \textit{Kitchen}.

\begin{table}[t]
    \begin{center}
    \caption{Candidate number statistics during the packing process.}
    \label{table:candidatenum}
    \setlength{\tabcolsep}{0.7em}
    \begin{tabular}{l|rr|rr|rr}
     & \multicolumn{2}{c|}{\textit{General}} & \multicolumn{2}{c|}{\textit{BlockOut}} & \multicolumn{2}{c}{\textit{Kitchen}} \\
    Method & Mean  & Max & Mean & Max & Mean  & Max  \\ 
    \noalign{\smallskip}
    \midrule
    \midrule
    \noalign{\smallskip}
    Ours & 195.7 & 868.0 & 51.0 & 172.0 & 140.3 & 605.0\\
    \end{tabular}
    \end{center}
    \end{table}

\subsection{\label{sec:actionDesign}Benefits of Candidate-based Packing Pattern}
One of our main contributions lies in action space design, i.e., \zhf{the candidate-based packing pattern for pruning sub-optimal actions.}
We highlight the effectiveness of our design in this section. 
We first compare 
with the resolution-complete action space, which discretizes the action space using a regular interval as done in~\cite{GoyalD20,HuangWZL23}. 
This action space can also be treated as a full geometry set including convex vertices, concave vertices, edges, and internal points of connected action regions, which would demonstrate the functionality of our generated candidates.
Then, we also 
compare 
with other dimension-reduction techniques 
below.

\begin{table*}[ht]
\begin{center}
\caption{Performance comparisons between various kinds of action space design along our five metrics.}
\label{table:actionSpace}
\setlength{\tabcolsep}{0.5em}
\begin{tabular}{l|crccc|crccc|crccc}
&  \multicolumn{5}{c|}{\textit{General}} & \multicolumn{5}{c|}{\textit{BlockOut}} & \multicolumn{5}{c}{\textit{Kitchen}} \\
Action Space Design & Uti. & Gap  & Var. & Num. & Time & Uti. & Gap  & Var. & Num. & Time  & Uti. & Gap  & Var. & Num. & Time \\ 
\noalign{\smallskip}
\midrule
\midrule
\noalign{\smallskip}
Resolution-Complete & 41.6\% & 6.5\% & 5.0 & 26.4 & 0.04 & 
55.7\% & 21.5\% & 3.4 & 27.7 & 0.02 & 
32.7\% & 16.6\% & 5.2 & 25.5 & 0.03 \\
Act in Line & 42.2\% & 5.2\% & 4.4 & 26.8 & \textbf{0.03} & 
60.3\% & 15.1\% & 2.7 & 29.9 & 0.02 & 
34.9\% & 11.0\% & \textbf{4.2} & 27.1 & 0.03 \\
Act on Orientation  & 38.6\% & 13.3\% & 5.2 & 24.6 & 0.03 & 
57.6\% & 18.9\% & 3.6 & 28.5 & 0.02  & 
32.1\% & 18.1\% & 5.6 & 25.1 & 0.03 \\
Act on Heuristics  & 40.1\% & 9.9\% & 6.0 & 25.5 & 0.03 & 
65.4\% & 7.9\% & 2.6 & 32.2 & 0.02 & 
32.8\% & 16.3\% & 5.6 & 25.4 & 0.03 \\
Ours & \textbf{44.5\%} & \textbf{0.0\%} & \textbf{3.4} & \textbf{27.7} & 0.04 & 
\textbf{71.0\%} & \textbf{0.0\%} & \textbf{1.7} & \textbf{34.8} & \textbf{0.02} & 
\textbf{39.2\%} & \textbf{0.0\%} & 4.5 & \textbf{29.4} &  \textbf{0.03} \\

\end{tabular}
\end{center}
\end{table*}

\textit{Act in Line.} The resolution-complete action space for packing is typically large and sensitive to the container size $S$ and the related intervals $\Delta$ for discretizing the space $SE(2) \times SO(1)$. We can get the  scale of this  discretized space $|\mathcal{A}|=\mathcal{O}(S^2\Delta^{-3})$. 
To sidestep such an enormous space that can grow explosively, we only require the robot to provide $l_x$ and $\theta$. The position $l_y$  can be determined via $\text{argmin}_{l_y}l_z(l_x,l_y)$ for given $\theta$, thus reducing the action space to $\mathbb{R} \times SO(1)$ with $|\mathcal{A}|=\mathcal{O}(S\Delta^{-2})$.

\textit{Act on Orientation.} Along the same line of thinking, we can further reduce 
the action space by only asking the robot to output $\theta$ and determine $l_x,l_y$ via $\text{argmin}_{l_x,l_y}l_z(l_x,l_y)$. Thus reducing the action space to $SO(1)$ with $|\mathcal{A}|=\mathcal{O}(\Delta^{-1})$.

\textit{Act on Heuristics.} Since heuristic methods have different packing preferences, a straightforward packing strategy is to select one of the heuristics for the object to be packed. We collect all heuristics mentioned in~Section~\ref{sec:comparisons} as a candidate action set $\textbf{m}$. The action space size, in this case, is $|\mathcal{A}|=|\textbf{m}|$.


For all alternatives mentioned above, we replace Q-values of invalid actions with negative infinity ($-\infty$) during  the decision-making process, and the subsequent performances are collated in~Table~\ref{table:actionSpace}. 
The widely adopted resolution-complete design of the action space~\cite{GoyalD20,HuangWZL23}, i.e. utilizing a brute-force discretization of 2D translation and rotation to form the action space, performs notably inferior to our method across all datasets. The reasoning behind this lies in  the large action space, which necessitates extensive RL exploration and exceeds feasible computational capacity.

Gaining advantages from dimension reduction,  Act-in-Line consistently outperforms resolution-complete actions. 
However, if we further reduce the action space to $SO(1)$, i.e., Act-on-Orientation, the decision flexibility decreases and correspondingly, the performance degrades. Unsurprisingly, Act-on-Heuristics surpasses the performance of each individual  heuristic reported in~Table~\ref{table:baselines}. However, the packing behavior of the learned policy is still restricted by basic heuristic components, and consequently, the improvements are limited. In comparison, our method stands out as the consistently best performer by reducing the
action space to  candidate actions of a fixed size $N$. 
This approach not only alleviates the learning burden but also enables compact object packing.


\subsection{\label{sec:generalization}Generalization Ability}
The generalization capability of learning-based methods, i.e., testing the trained policies with a problem emitter differing from the training one, has always been a concern. Here we demonstrate our generalization proficiency  with two experiments from practical perspectives.

\paragraph{Generalization on Noisy Point Clouds}
We utilize a point cloud to represent the incoming object, which is then re-sampled randomly from  $P$. To exhibit our generalization capability, we apply Gaussian noises $d \cdot N(0,\sigma^2)$ to the re-sampled $P$.
Here $\sigma$ represents the standard deviation and $d$ is the diagonal length of $P$'s AABB. Then we generalize the policies trained on $\sigma = 0$ to $\sigma = 1\%-10\%$.

\paragraph{Generalization on Unseen Shapes}
To examine our method's performance on unseen shapes, we randomly exclude $20\%$ of shapes from each dataset for training and we test the trained policies on the complete datasets afterwards. For the \textit{Kitchen} dataset, this shape removal is executed  by each category. 

\begin{table}[!t]
\begin{center}
\caption{The performance of our method when observation is corrupted by point cloud noises of different amplitude $\sigma$.}
\label{table:noise}
\setlength{\tabcolsep}{0.3em}
\begin{tabular}{l|ccc|ccc|ccc}
 &  \multicolumn{3}{c|}{\textit{General}} & \multicolumn{3}{c|}{\textit{BlockOut}} & \multicolumn{3}{c}{\textit{Kitchen}} \\
Noise & Uti. & Var. & Num. & Uti.  & Var. & Num. & Uti.  & Var. & Num.  \\ 
\noalign{\smallskip}
\midrule
\midrule
\noalign{\smallskip}
$\sigma=0$ & 44.5\%  & 3.4 & 27.7 &
71.0\% & 1.7 & 34.8 &
39.2\% & 4.5 & 29.4 \\
$\sigma=1\%$ & 44.5\%  & 3.5 & 27.7 &
70.7\% & 1.7 & 34.7 &
38.8\% & 4.5 & 29.0 \\
$\sigma=3\%$ & 44.4\%  & 3.4 & 27.6 &
70.7\% & 1.7 & 34.7 &
38.9\% & 4.3 & 29.1 \\
$\sigma=5\%$ & 44.3\%  & 3.4 & 27.6 &
70.5\% & 1.8 & 34.6 &
38.5\% & 4.5 & 28.9 \\
$\sigma=10\%$ & 43.6\%  & 3.6 & 27.2 &
70.4\% & 1.7 & 34.6 &
37.3\% & 48.2 & 28.1 \\
\end{tabular}
\end{center}
\end{table}

\begin{table}[t]
\begin{center}
\caption{Performance of our method trained with partial object datasets. We also provide test results from  randomly initialized  policies  as a baseline.}
\label{table:unseen}
\setlength{\tabcolsep}{0.26em}
\begin{tabular}{c|c|ccc|ccc|ccc}
         & & \multicolumn{3}{c|}{\textit{General}} & \multicolumn{3}{c|}{\textit{BlockOut}} & \multicolumn{3}{c}{\textit{Kitchen}} \\
Train& Test & Uti. & Var. & Num. & Uti.  & Var. & Num. & Uti.  & Var. & Num.  \\ 
\noalign{\smallskip}
\midrule
\midrule
\noalign{\smallskip}
- & Full & 32.9\%  & 4.6 & 20.4 &
42.1\% & 3.5 & 20.2 &
25.8\% & 5.5 & 18.9 \\ 
Full & Full & 44.5\%  & 3.4 & 27.7 &
71.0\% & 1.7 & 34.8 &
39.2\% & 4.5 & 29.4 \\
Part & Full & 44.3\%  & 3.5 & 27.6 &
70.7\% & 1.4 & 34.7 &
39.1\% & 4.5 & 29.2 \\
\end{tabular}
\end{center}
\end{table}


The generalization outcomes on noisy point clouds and unseen shapes are presented in~Table~\ref{table:noise} and~Table~\ref{table:unseen}, respectively. Our method retains its performance across diverse amplitude of point cloud noises and continues to function effectively  even under  a significant noise disturbance of $\sigma = 10\%$. The policies trained with only part of each dataset  exhibit the ability to adapt well to the full dataset, accompanied by a negligible decrease in performance. Note that, even the worst generalization performance in~Table~\ref{table:noise} and~Table~\ref{table:unseen} still clearly outperforms the best heuristic performance as reported in~Table~\ref{table:baselines},  thus showcasing the robustness of our learned policies. Additionally, we have conducted experiments to test the applicability of trained policies across different datasets, as detailed in~Section~\ref{sec:betweenDatasets}.

\subsection{\label{sec:buffeResult}Performance on Buffered Packing Scenario}

\begin{table*}[t!]
\begin{center}
\caption{The buffered packing problem can be well solved by training an additional object-ordering policy $\pi_s$ cooperating with the placement policy $\pi$. The online case with $K=1$ can be treated as a baseline, where no object-ordering policy exists and the objects are randomly dispatched.}
\label{table:resultBuffer}
\setlength{\tabcolsep}{0.4em}
\begin{tabular}{c|c|c|crccc|crccc|crccc}
& & &  \multicolumn{5}{c|}{\textit{General}} & \multicolumn{5}{c|}{\textit{BlockOut}} & \multicolumn{5}{c}{\textit{Kitchen}} \\
 Train & Test &Method & Uti. & Gap  & Var. & Num. & Time & Uti. & Gap  & Var. & Num. & Time  & Uti. & Gap  & Var. & Num. & Time \\ 
\noalign{\smallskip}
\midrule
\midrule
\noalign{\smallskip}
$K=1$&$K=1$ & $\pi$ &  44.5\% & \multicolumn{1}{c}{-} & 3.4 & 27.7  & 0.04  & 
71.0\% & \multicolumn{1}{c}{-} & 1.7 & 34.8 & 0.02 & 
39.2\% & \multicolumn{1}{c}{-} & 4.5 & 29.4 &  0.03 \\
\midrule
&&LFSS \& $\pi$   & 44.3\% & 2.2\% & 3.5 & 25.8 & 0.07 & 
70.8\% & 1.3\% & \textbf{1.5} & 33.4 &  0.04 & 
39.7\% & 10.4\% & \textbf{4.7} & 28.3 &  0.06\\
$K=10$&$K=3$&$\pi_s$ \& BLBF   & 42.0\% & 7.3\% & 3.9 & 27.8 &  \textbf{0.03} & 
66.9\% & 6.7\% & 2.7 & 33.2 &  \textbf{0.02} & 
38.1\% & 14.0\% & 4.8 & 29.5 &  \textbf{0.03} \\
&& $\pi_s$ \& $\pi$  & \textbf{45.3\%} & \textbf{0.0\%} & \textbf{3.4} & \textbf{28.9} & 0.07 & 
\textbf{71.7\%} & \textbf{0.0\%} & 2.4 & \textbf{35.5} & 0.04 & 
\textbf{44.3\%} & \textbf{0.0\%} & 5.5 & \textbf{34.1} & 0.07 \\
\midrule
&& LFSS \& $\pi$   & 44.4\% & 6.5\% & 3.6 & 24.1 & 0.07 & 
70.5\% & 5.7\% & 1.5 & 32.1 &  0.04 & 
40.1\% & 17.8\% & \textbf{4.9} & 27.4 &  0.06 \\
$K=10$&$K=5$&$\pi_s$ \& BLBF   & 45.3\% & 4.6\% & 4.1 & 30.3 &  \textbf{0.04} & 
70.9\% & 5.2\% & 2.1 & 35.3 &  \textbf{0.02} & 
42.3\% & 13.3\% & 5.0 & 32.9 &  \textbf{0.03} \\
&&$\pi_s$ \& $\pi$ & \textbf{47.5\%} & \textbf{0.0\%} & \textbf{3.1} & \textbf{31.4} & 0.07 & \textbf{74.8\%} & \textbf{0.0\%} & \textbf{1.5} & \textbf{37.2} & 0.05 & \textbf{48.8\%} & \textbf{0.0\%} & 5.5 & \textbf{37.8} & 0.07  \\
\midrule
&& LFSS \& $\pi$    &  45.7\% & 18.0\% & 3.4 & 20.9 &  0.07 & 
69.3\% & 11.2\% & 1.8 & 29.6 &  0.04 & 
41.8\% & 27.7\% & \textbf{5.6} & 26.1 &  0.07 \\
$K=10$&$K=10$& $\pi_s$ \& BLBF & 49.6\% & 11.0\% & 4.0 & 32.9 &  \textbf{0.04} & 
74.8\% & 4.1\% & 2.0 & 36.9 &  \textbf{0.02} & 
50.5\% & 12.6\% & 5.9 & 38.7 &  \textbf{0.03} \\
&&$\pi_s$ \& $\pi$ & \textbf{55.7\%} & \textbf{0.0\%} & \textbf{2.3} & \textbf{38.5} & 0.07 & \textbf{78.0\%} & \textbf{0.0\%} & \textbf{0.9} & \textbf{38.5} & 0.05 & \textbf{57.8\%} & \textbf{0.0\%} & 6.4 & \textbf{44.2} & 0.07  \\

\end{tabular}
\end{center}
\end{table*}

\begin{figure}[t!]
    \begin{center}
    \centerline{\includegraphics[width=0.5
    \textwidth]{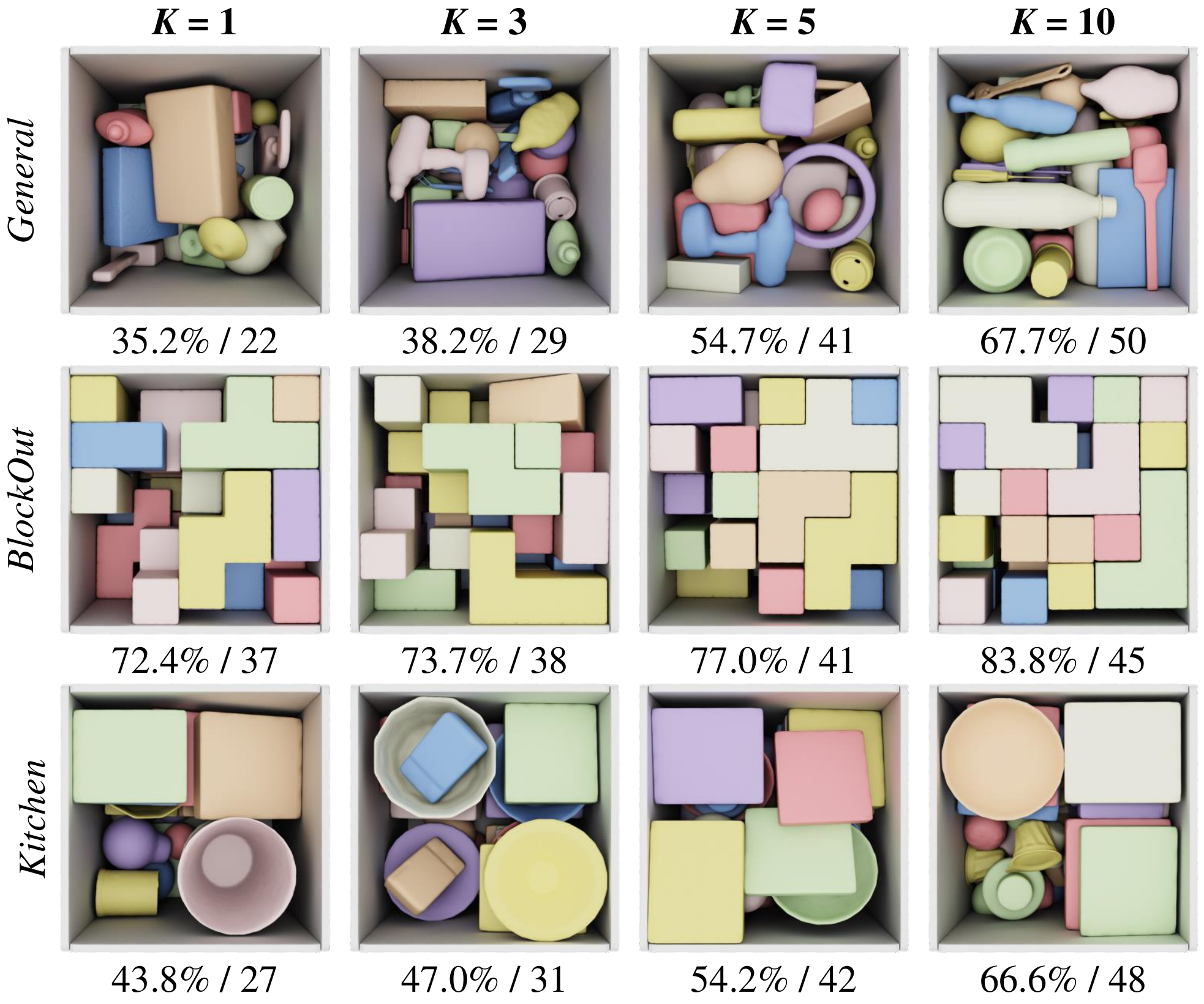}}
    \caption{
        Visualization results of our method on buffered packing scenarios with various $K$. The larger buffer provides more flexibility for the object-ordering policy $\pi_s$ and results in a more dense packing.
    }
\label{fig:visualBuffer}
\vspace{-20pt}
\end{center}
\end{figure}

We demonstrate that our method can naturally address the buffered packing problem by simply introducing an  object-ordering policy $\pi_s$,  to cooperate with the existing placement policy $\pi$. 
To ascertain that both $\pi_s$ and $\pi$ are necessary components, we compare them with the systematically corresponding heuristic baselines proposed by~\citet{GoyalD20}. 
Alongside the placement heuristic BLBF mentioned above, \citet{GoyalD20} also suggest an LFSS method for ordering objects in a largest-volume-first preference. We amalgamate LFSS with $\pi$, and combine $\pi_s$ with BLBF to serve as our baselines. All learning-based policies are trained with buffer size $K=10$. Further, to exemplify the generalization capability, we test the trained policies across various values of $K$, as illustrated  in~Fig.~\ref{fig:visualBuffer} and summarized in~Table~\ref{table:resultBuffer}.

Our comparisons show that introducing an object-ordering policy $\pi_s$ for buffered packing scenarios with $K=10$ significantly improves the packing performance compared with the strictly online case, where $\pi_s$ does not exist and $K=1$. The jointly trained policies $\pi_s$ and $\pi$ outperform the other two combination alternatives
by a considerable margin on each dataset. 
This advantage is sustained when we generalize the trained policies to buffered packing scenarios with $K=3$ and $K=5$. 
More packing results of our method on buffered packing are visualized in Appendix~\ref{sec:appendixB}.

\begin{figure}[t!]
    \begin{center}
    \centerline{\includegraphics[width=0.5
    \textwidth]{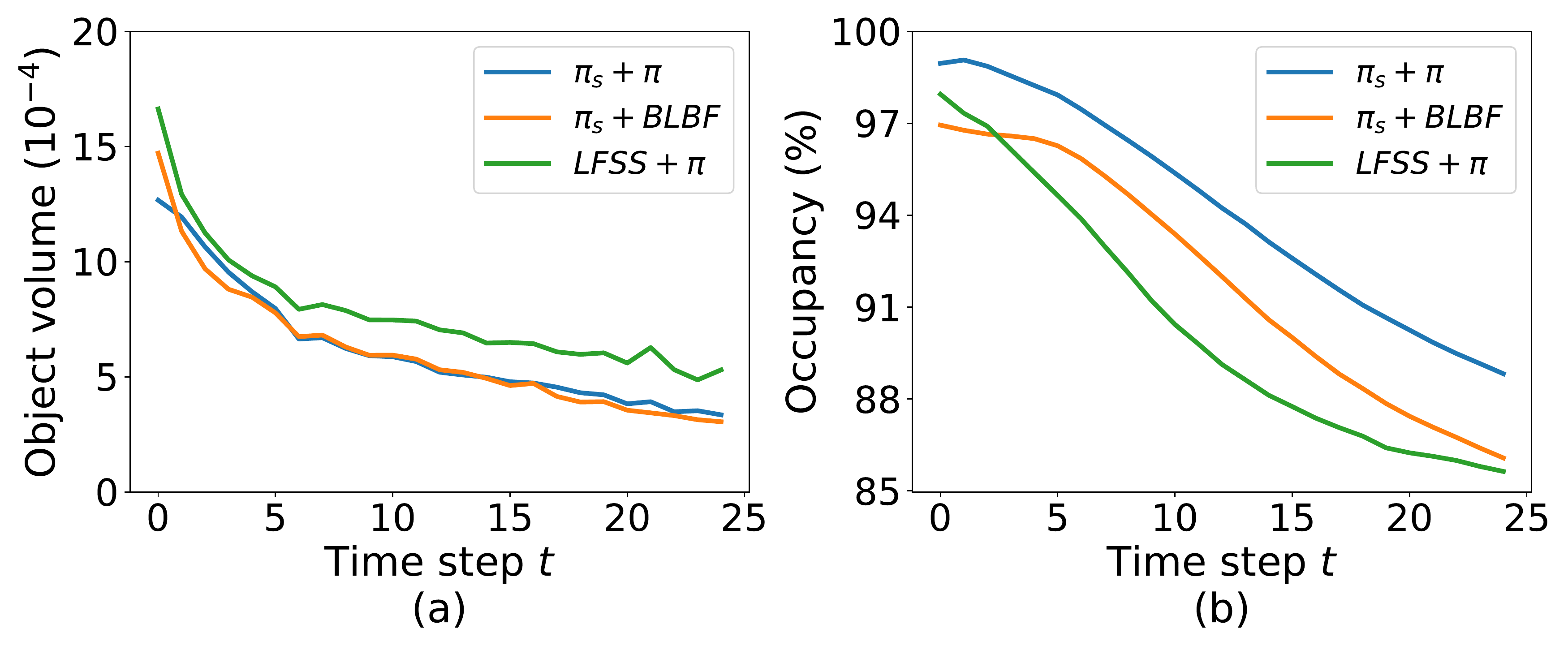}}
    \vspace{-6pt}
    \caption{
        Packing behavior analysis on the $General$ dataset with $K = 10$. The policies $\pi_s$ and $\pi$ both contribute to better  utilization of occupied spaces.
    }
\label{fig:sizeRatio}
\end{center}
\vspace{-20pt}
\end{figure}

To figure out what the two policies learned, we calculate the average volume of objects chosen by $\pi_s$ during the packing process and visualize it in~Fig.~\ref{fig:sizeRatio}a.
We can see that $\pi_s$ automatically learns a strategy that selects objects from large to small like LFSS.
We visualize a metric  $\sum |G_{i(t)}| / V_f$ in~Fig.~\ref{fig:sizeRatio}b to reflect space occupancy, where $V_f$ is the volume below the up surface composed of packed objects.
We can see that $\pi_s$ can select more suitable items to  keep higher occupancy than LFSS so that occupied space $V_f$ is better utilized.
Also, we can get that the learnable $\pi$ also contributes better packing by comparing the occupancy between  $\pi_s + \text{BLBF}$ and $\pi_s + \pi$.

\section{\label{sec:conclusion}Conclusion and Future Work}
We investigate problem setups and solution techniques for learning online packing skills for irregular 3D shapes.
We propose a learning-based  method to successfully pack objects with complex 3D shapes at real-time rates, while  taking physics dynamics and constraints of a placement into account.
Our theoretically-provable candidate generation algorithm prunes sub-optimal actions and forms a set of placements for a learnable policy, leading to high-quality packing plans. Equipped with asynchronous RL acceleration techniques and a data preparation process of simulation-ready training sequences, a mature packing policy can be trained within 48 hours in a physically realistic environment. 
Through evaluations on a variety of real-life object datasets, our performance beats state-of-the-art baselines in terms of both packing utilities and the number of packed objects. Our method can also be naturally extended to solve buffered packing problems by only introducing an additional object-ordering policy. 

Our results shed light on many other packing-related problems in the graphics community including UV generation and 3D printing. 
We also release our datasets and source code to support further research in this direction.
Our work has two limitations. Firstly, we model the general irregular shapes as rigid bodies and neglect their material.
Secondly, we have only experimented with our method in simulation, where we always assume robots can successfully pick up objects on the conveyor belt using a sucker-type gripper. Future works could involve experiments on hardware platforms, grasp planning for real-world objects, and handling failure cases.
Finally, we are also interested in introducing the stress metric~\citep{9196938} for better placing fragile objects, as well as  incorporating  gripper feasibility into the packing decision. Exploiting deformation during planning to achieve tighter packing~\cite{YinVK21} is also an interesting direction.

\section{ACKNOWLEDGEMENTS}

The authors acknowledge the anonymous reviewers for their insightful comments and valuable suggestions. 
Thanks are also extended  to  Yin Yang, Qijin She, Juzhan Xu,  Lintao Zheng, and Jun Li for their helpful discussions.
Hang thanks Tong Zhang with heartfelt appreciation for her support, understanding, and encouragement.
This work is supported by the National Key Research and Development Program of China (2018AAA0102200), and the National Natural Science Foundation of China (62132021, 62102435).

\bibliographystyle{ACM-Reference-Format}
\bibliography{IRBPP}
\appendix
\clearpage  
\section*{Appendix}
\setcounter{section}{0}



\section{\label{sec:appendixA}Pipeline for Data Preparation}

We outline our specific data preparation pipeline in~Algorithm~\ref{alg:data}, where a polygonal mesh is transformed into several watertight convex decompositions, and its planar-stable poses are provided.  
Considering that some stable poses are rotation-symmetric, we propose~Algorithm~\ref{alg:poses} to remove redundant poses to avoid unbalanced shape distribution.

\begin{algorithm} 
    \caption{Object Data Preparation}       
    \label{alg:data} 
    \begin{algorithmic}[1]
      \STATE \textbf{Input:} 
      A polygonal mesh $M$ obtained by arbitrary approaches.
      \IF{$M$ is not watertight}
      \STATE Recondstuct $M$~\citep{StutzG20} to ensure watertightness.
      \ENDIF
      \STATE Compute all possible planar-stable poses $\textbf{P}$~\citep{GoldbergMZCCC99} of $M$.
      \STATE Decompose $M$ into several convex parts~\citep{mamou2016volumetric} if non-convex. This step also simplifies $M$ with fewer vertices, which benefits collision detections for simulation.
      \STATE Place the decomposed $M$ in the simulator~\cite{coumans2016pybullet} with poses in $\textbf{P}$. Remove poses which are factually unstable.   
      \STATE Remove redundant rotation-symmetric poses in $\textbf{P}$ with~Algorithm~\ref{alg:poses}.
      \STATE \textbf{Return} Convex decompositions of $M$ and $\textbf{P}$.
    \end{algorithmic}
  \end{algorithm}

\begin{algorithm} 
    \caption{Remove Redundant Planar-Stable Poses}       
    \label{alg:poses} 
    \begin{algorithmic}[1]
      \STATE \textbf{Input:} 
      A watertight mesh $M$, poses $\textbf{P}$, and a percentage constant $c$.
      \STATE Valid pose set $\textbf{P}_v \gets \emptyset$,  occupancy label set $\textbf{L} \gets \emptyset$. 
      \FOR{each pose $p \in \textbf{P}$}
      \STATE Place $M$ with $p$ and rotate $M$ vertically with the condition $s_x \le s_y$ and $m_x \le m_y$ until the smallest AABB volume is reached.  Here $s$ is the AABB size of $M$ and $m$ the mass center.  Rewrite $p\in\textbf{P}$.
      \ENDFOR
      \STATE Calculate the maximal AABB size after rotation. Denoted the size maximum along each dimension $d_x, d_y$, and $d_z$.
      \STATE Voxelize the space in the range $[0, d_x] \times [0,d_y] \times [0, d_z]$. 
      \FOR{each rotated pose $p\in\textbf{P}$}
      \STATE Move $M$ with the FLB corner of its AABB aligned to the origin. 
      \STATE Check if each space voxel is occupied by $M$ with pose $p$. 
      Denote the occupancy of all voxels as $l_p$.  
      \IF{no $l \in \textbf{L} $ make $\sum f_x(l, l_p)/\sum l_p \le c$, where $f_x$ is the xor function}
      \STATE Insert $p$ into $\textbf{P}_v$ and insert $l_p$ into $\textbf{L}$.
      \ENDIF
      \ENDFOR
      \STATE \textbf{Return} $\textbf{P}_v$.
    \end{algorithmic}
  \end{algorithm}


\section{\label{sec:appendixB}More Experimental Results}
\subsection{\label{sec:resolutionAbaltion}Effects of Experimental Parameters}

Our packing experiment setup involves a set of parameters for finding candidates and describing packing observations. Here we refine these parameters to make ablation and study their effect on the final packing performance. 
We conduct this experiment on our main dataset \textit{General}.
We double the number of points sampled from object surfaces to $2048$ and the candidate number $N$ to $1000$. For the intervals $\Delta_h, \Delta_\theta, \Delta_z, $ and $\Delta_g$ used to find candidates, we halve them to investigate whether finer action can lead to better performance. 

\begin{table}[ht!]
    \begin{center}
    \caption{This table studies the effects of \zhf{finer} experimental parameters.}
    \label{table:ablation}
    \setlength{\tabcolsep}{0.6em}
    \begin{tabular}{l|cccc}
    Parameters & Uti.  & Var. & Num. & Time  \\ 
    \noalign{\smallskip}
    \midrule
    \midrule
    \noalign{\smallskip}
    Baseline & 44.5\% & 3.4 & 27.7 & 0.04 \\
    Double Point Cloud Number & 42.2\% & 2.1 & 26.3 & 0.04 \\ 
    Double Candidate Actions $N$ & 42.6\% & 2.0 & 26.5 & 0.04 \\
    Half Heightmap Interval $\Delta_h$ & 43.8\% & 2.3 & 27.0 & 0.13 \\ 
    Half Rotation Interval $\Delta_\theta$  & 42.1\% & 2.1 & 26.3 & 0.07 \\ 
    Half Height Interval $\Delta_z$ & 42.2\% & 2.3 & 26.3 & 0.05 \\ 
    Half Pixel Interval $\Delta_g$ & 44.8\% & 2.3 & 27.8 & 0.13 \\ 
    \end{tabular}
    \vspace{-10pt}
    \end{center}
    \end{table}

We summarize the test results in~Table~\ref{table:ablation}. 
Halving the pixel interval $\Delta_g$ used to sample grid points has improvement to the packing performance, but it also substantially increases the time for decision-making.
Finer tuning of other parameters 
no longer has a clear impact and may increase the computational overheads.
We choose a set of efficient and effective parameters to do our main experiments.

\subsection{\label{sec:trainingAbaltion}Ablations on RL Training}

Here we do ablation studies to demonstrate the efficacy of our asynchronous RL training. We provide policies trained with the vanilla Rainbow as a baseline. We remove the distributed learner and the non-blocking actor equipped with batched simulation to illustrate their effect. We also report the performance without transforming the point cloud observation to a canonical configuration. All these policies are trained within $48$ hours for fairness. From the test results summarized in~Table~\ref{table:architecture}, we can see that our asynchronous training achieves the best performance. Removing the canonical transform affects the data efficiency and lowers the packing utility.

\begin{table}[t]
\begin{center}
\caption{Advantages of our asynchronous reinforcement learning training fashion. This experiment is conducted on the \textit{General} dataset.
}
\label{table:architecture}
\setlength{\tabcolsep}{0.6em}
\begin{tabular}{l|cccc}
Training Variant &  Uti. & Gap  & Var. & Num. \\ 
\noalign{\smallskip}
\midrule
\midrule
\noalign{\smallskip}
The Vanilla Rainbow  & 41.3\% & 7.2\% & \textbf{2.3} & 25.7 \\ 
No Distributed Learner & 42.9\% & 3.6\% & 3.3 & 26.8 \\   
No Batched Simulation & 41.3\% & 7.2\% & 3.3 & 26.0 \\  
No Canonical Transform & 40.2\% & 9.7\% & 3.0 & 25.3 \\ 
Our Asynchronous Training & \textbf{44.5\%} & \textbf{0.0\%} & 3.4 & \textbf{27.7} \\  
\end{tabular}
\vspace{-10pt}
\end{center}
\end{table}

\subsection{Results on Shapes from ABC}
\begin{figure}[t]
    \begin{center}
    \centerline{\includegraphics[width=0.5
    \textwidth]{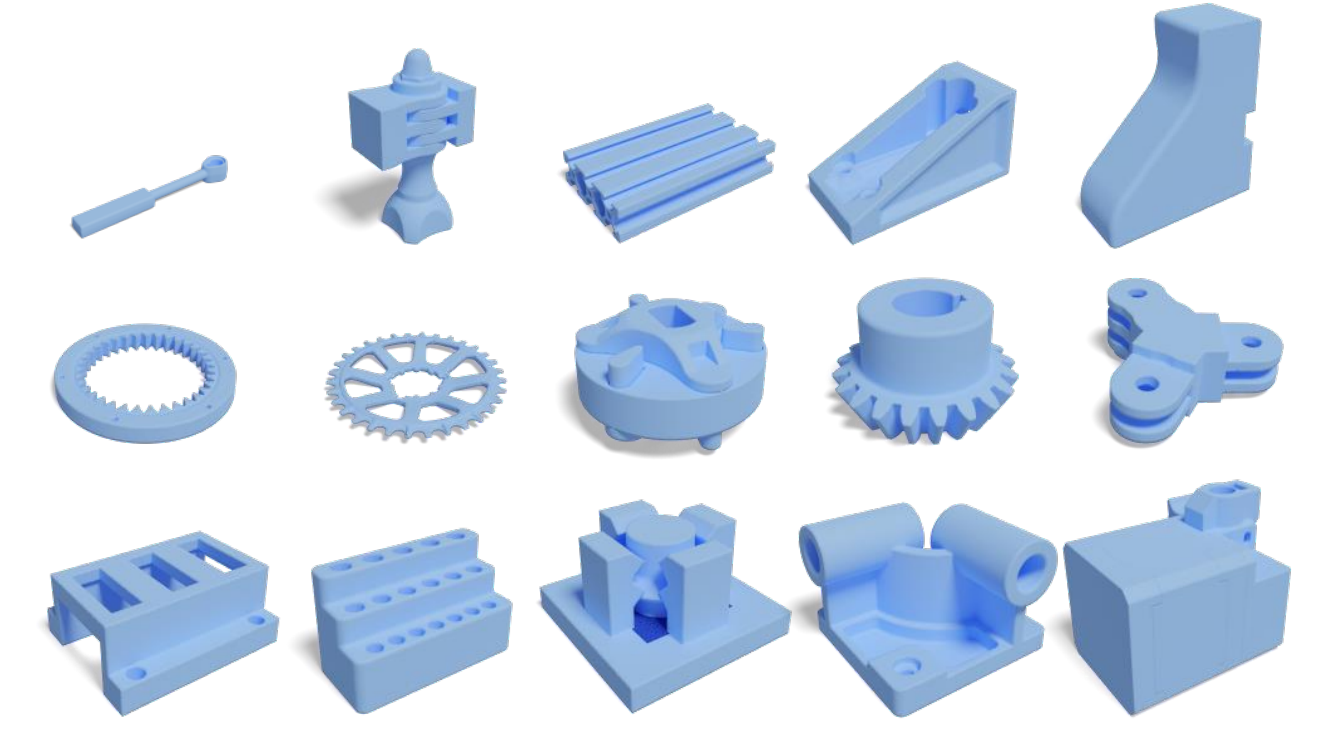}}
    \caption{
        Mechanical shapes selected from the ABC dataset.
        These shapes are scaled with their maximal AABB size equal to 1 for clarity.
    }
\label{fig:abc}
\end{center}
\vspace{-20pt}
\end{figure}

We test our method on shapes collected from the ABC dataset~\cite{KochMJWABAZP19}. 
These shapes are mostly complex mechanical parts with distinct characteristics, as shown in ~Fig.~\ref{fig:abc}. 
Totally 136 industrial shapes with 440 planar-stable poses are collected. 
We train an online packing policy and compare it with the existing heuristic competitors to demonstrate the superiority of our method. We also train the object-ordering and the placement policy pairs for solving buffered packing problems. We train these  policy pairs with buffer size $K = 10$ and generalize them to buffered packing scenarios with $K=3$ and $K=5$.
We report these results in ~Table~\ref{table:abcresult}.

\begin{table}[t]
\begin{center}
\caption{Packing performance on the shapes for the ABC dataset.}
\label{table:abcresult}
\setlength{\tabcolsep}{0.9em}
\begin{tabular}{l|crccc}
Method & Uti. & Gap  & Var. & Num. & Time  \\ 
\noalign{\smallskip}
\midrule
\midrule
\noalign{\smallskip}
Random & 
22.0\% & 35.3\% & 2.7 & 14.2 & 0.02 \\
Random $\pi$ & 
23.6\% & 30.6\% & 2.3 & 15.2 & 0.03 \\
\midrule
FF~\shortcite{Falkenauer96} & 
25.0\% & 26.5\% & 2.5 & 16.0 &   0.03 \\
MTPE~\shortcite{LiuLCY15} & 
29.3\% & 13.8\% & 1.9  & 18.7 &   \textbf{0.03} \\
HM~\shortcite{WangH19a} & 27.4\% & 19.4\% & 2.0  & 17.5 &   0.03 \\
BLBF~\shortcite{GoyalD20} & 28.7\% & 15.6\% &  1.9 & 18.3  &   0.03 \\
Ours & 
\textbf{34.0\%} & \textbf{0.0\%} & \textbf{1.9} & \textbf{22.6} & 0.04 \\
\midrule
Ours (K = 3) & 
35.3\% & \multicolumn{1}{c}{-} & 1.7 & 23.5 & 0.06 \\
Ours (K = 5) & 
37.4\% & \multicolumn{1}{c}{-} & 1.6 & 24.5 & 0.06 \\
Ours (K = 10) & 
41.0\% & \multicolumn{1}{c}{-} & 1.7 & 25.5 & 0.06 \\
\end{tabular}
\vspace{-10pt}
\end{center}
\end{table}

\subsection{Generalization across Datasets\label{sec:betweenDatasets}}
We test generalization across datasets, that is, crossly test trained policies on other datasets. 
Since part of the shapes is shared between $General$ and $Kitchen$, we conduct this experiment among $General$,  shapes from the ABC dataset, and $BlockOut$, with 483, 136, and 8 shapes respectively. These results are summarized in ~Table~\ref{table:generation}.
When being transferred to a new dataset, the trained policies can still show decision-making ability more competitive than heuristics.

\begin{table}[t]
    \begin{center}
    \caption{Generalize the trained policies to new datasets. The random performance and the best heuristic performance are also listed here. 
    }
    \label{table:generation}
    \setlength{\tabcolsep}{0.26em}
    \begin{tabular}{l|ccc|ccc|ccc}
    & \multicolumn{3}{c|}{\textit{General}} & \multicolumn{3}{c|}{\textit{ABC}} & \multicolumn{3}{c}{\textit{BlockOut}} \\
    Train & Uti. & Var. & Num. & Uti.  & Var. & Num. & Uti.  & Var. & Num.  \\ 
    \noalign{\smallskip}
    \midrule
    \midrule
    \noalign{\smallskip}
    \textit{General} & 44.5\%  & 3.4 & 27.7 &
    33.7\% & 1.7 & 22.4 & 
    67.5\% & 1.6 & 33.2 \\
    \textit{ABC} & 43.3\%  & 4.1 & 27.2 &
    34.0\% & 1.9 & 22.6 &
    66.1\% & 1.4 & 32.4 \\
    \textit{BlockOut} & 38.0\%  & 3.7 & 24.2 &
    29.7\% & 1.4 & 20.0 &
    71.0\% & 1.7 & 34.8 \\
    \midrule
    \textit{Random} & 31.6\%  & 5.8 & 19.6 &
    22.0\% & 2.7 & 14.2 & 
    37.1\% & 3.4 & 17.8 \\
    \textit{Heuristic} & 37.3\%  & 4.6 & 22.7 &
    29.3\% & 1.9 & 18.7 & 
    61.9\% & 4.1 & 29.5 \\
    \end{tabular}
    \vspace{-10pt}
    \end{center}
    \end{table}

\begin{figure}[t!]
    \begin{center}
    \centerline{\includegraphics[width=0.48
    \textwidth]{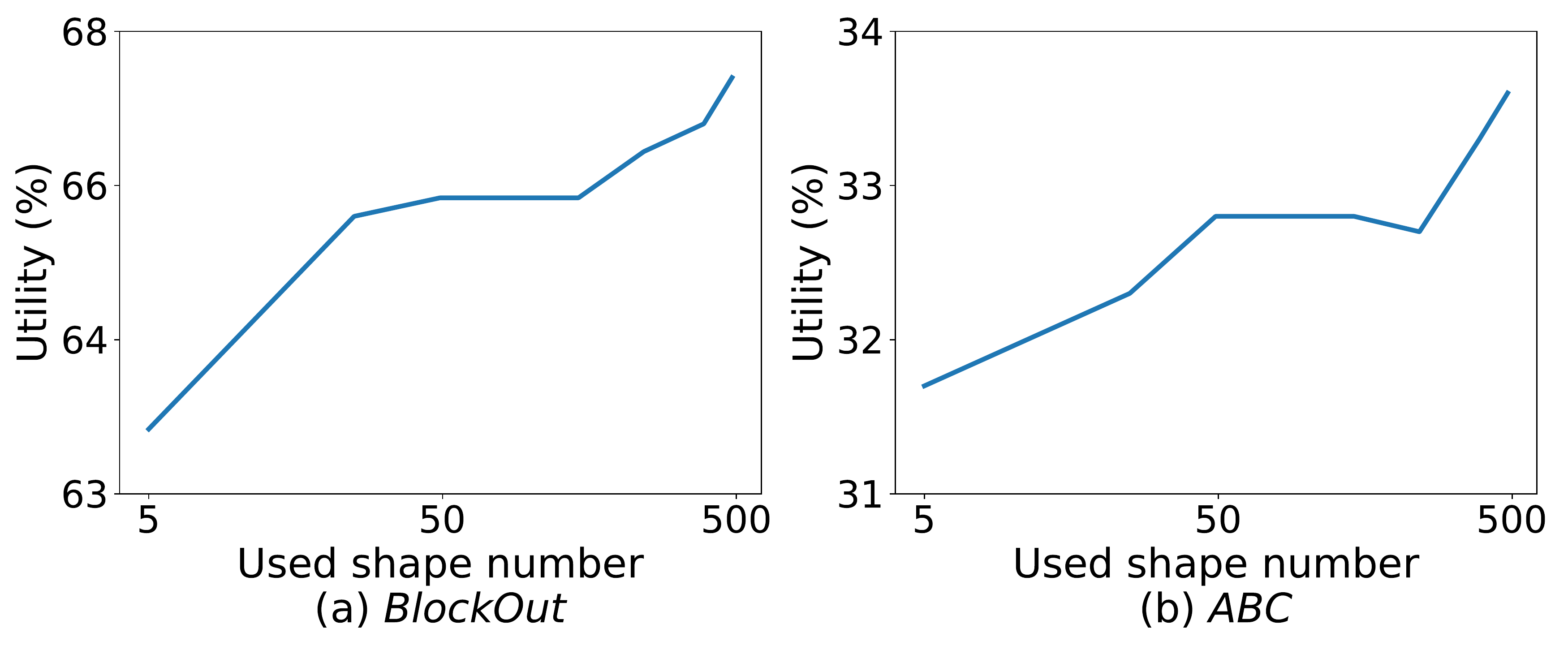}}
    \vspace{-6pt}
    \caption{
        Training policies with more shapes from $General$ benefits the performance when being generalized to the other two datasets. }
\label{fig:dataVariety}
\end{center}
\vspace{-10pt}
\end{figure}

We note that policies trained on datasets with more variety of shapes tend to perform better on one out-of-distribution dataset. To confirm this, we train policies with  different numbers of shapes from $General$ and test them on ABC and $BlockOut$.
Results are visualized in~Fig.~\ref{fig:dataVariety}. We can see that rich training shapes help policies transfer to the other two datasets. 
We recommend that users increase shape variety to enable better performance on out-of-distribution shapes.

\subsection{Product Utility \label{sec:productutility}}
We evaluate packing results with a new product utility metric, which is defined as the initial bounding box volume of all placed items divided by the container volume:
\begin{align}
\label{eq:productutility}
\sum_{G_i \subset C} |G_i|/|C|, 
\end{align}
where $|C|$ is the container volume. This metric is widely used in the manufacturing industry to reveal the connection between irregular and bin packing. We calculate this metric on all datasets and present the results in~Table~\ref{table:productutility}.
\begin{table}[t]
    \begin{center}
    \caption{Product utility on each dataset.}
    \label{table:productutility}
    \setlength{\tabcolsep}{0.9em}
    \begin{tabular}{l|rrrr}
    Method & \textit{General}  & \textit{BlockOut} & \textit{Kitchen} & \textit{ABC} \\ 
    \noalign{\smallskip}
    \midrule
    \midrule
    \noalign{\smallskip}
    Random & 46.5\% & 47.0\% & 48.1\% & 43.7\% \\
    Random $\pi$ & 48.3\% & 53.4\% & 54.5\% & 46.8\% \\
    \midrule
    FF~\shortcite{Falkenauer96} & 53.2\% & 43.0\% & 59.0\% & 49.6\% \\
    MTPE~\shortcite{LiuLCY15} & 54.8\% & 58.4\% & 69.3\% & 58.0\% \\
    HM~\shortcite{WangH19a} & 52.6\% & 59.5\% & 70.0\% & 53.9\% \\
    BLBF~\shortcite{GoyalD20} & 53.9\% & 61.9\% & 68.7\% & 57.2\% \\
    Ours & 65.1\% & 90.1\% & 85.4\% & 67.1\% \\
    \midrule
    Ours (K = 3)  & 70.4\% & 90.7\% & 88.3\% & 68.6\% \\
    Ours (K = 5)  & 75.6\% & 94.5\% & 91.7\% & 70.9\% \\
    Ours (K = 10) & 82.7\% & 98.6\% & 97.4\% & 73.9\% \\
    \end{tabular}
    \vspace{-10pt}
    \end{center}
    \end{table}

\vspace{10pt}
\section{\label{sec:moreVisual}More Visualized Results}
\begin{figure*}[ht]
\begin{center}
\centerline{\includegraphics[width=0.92\textwidth]{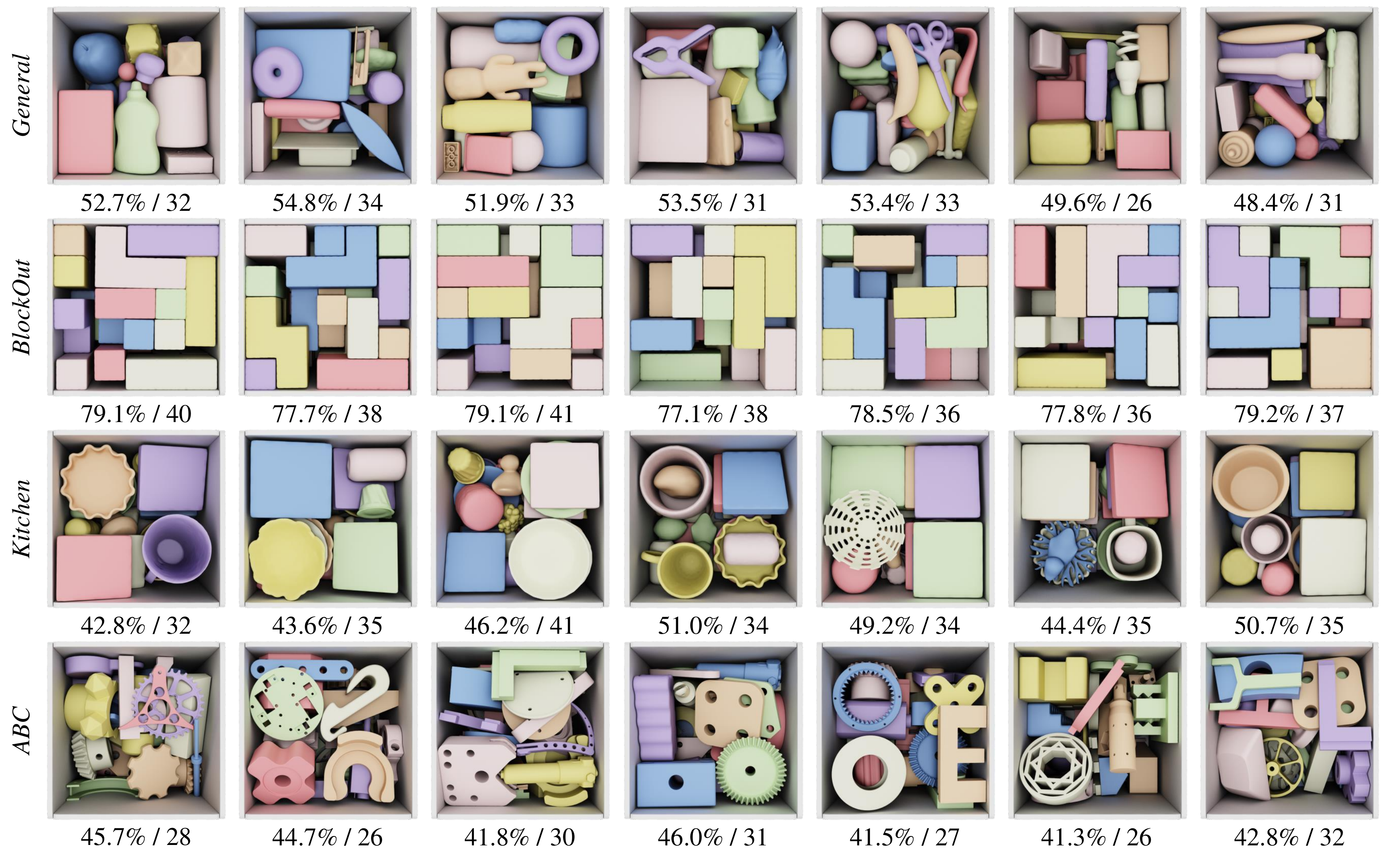}}
\vspace{-10pt}
\caption{\label{fig:galleryOnline} Results generated by our online packing policies.
Their utility and number of packed objects are labeled. }
\end{center}
\vskip -0.2in
\end{figure*}

\begin{figure*}[ht]
\begin{center}
\centerline{\includegraphics[width=0.92\textwidth]{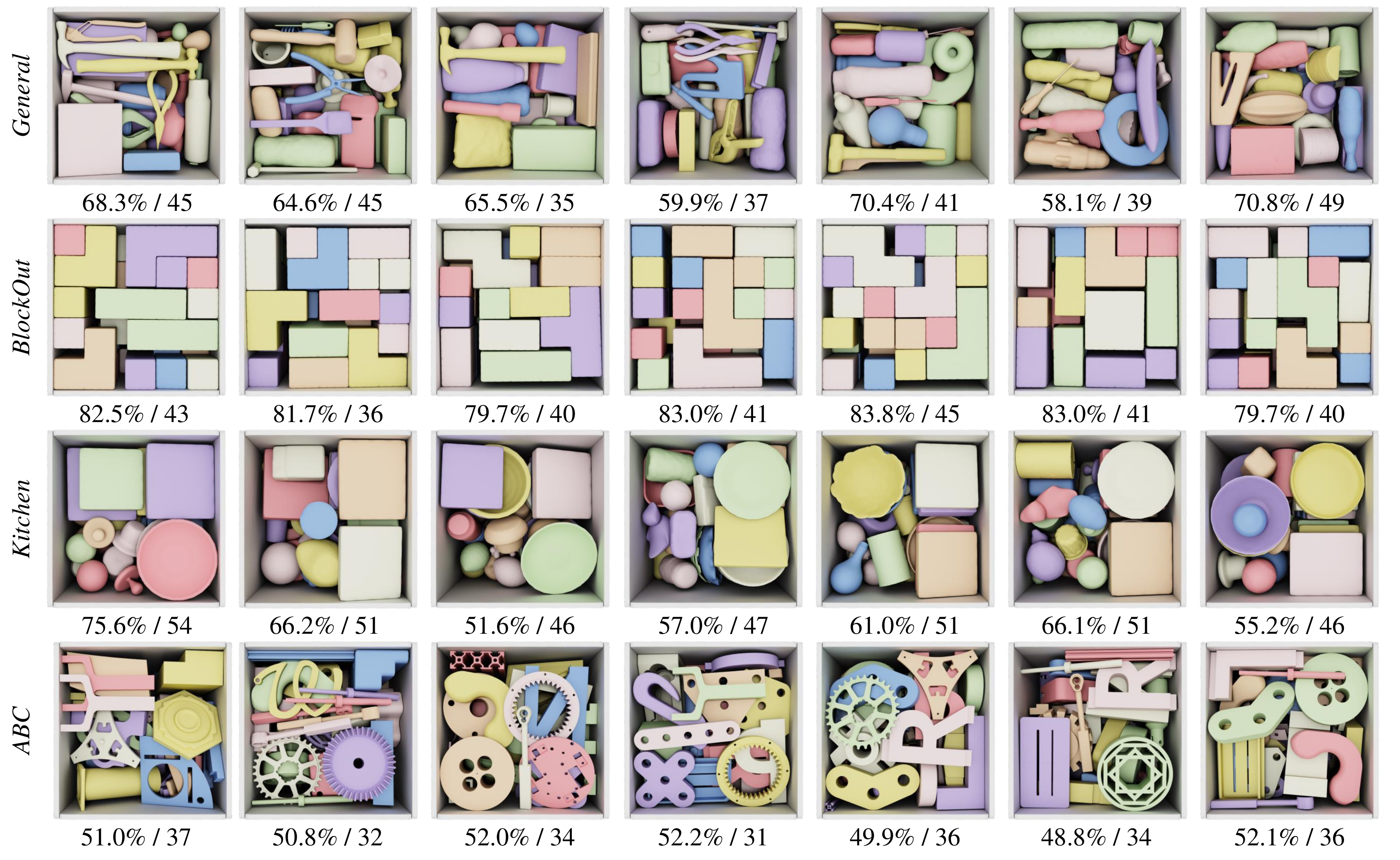}}
\vspace{-10pt}
\caption{Results generated by our buffered packing policies. These policies are trained and tested with a buffer size $K=10$.}
\label{fig:galleryBuffer1}
\end{center}
\vskip -0.2in
\end{figure*}

\begin{figure*}[ht]
\begin{center}
\centerline{\includegraphics[width=0.92\textwidth]{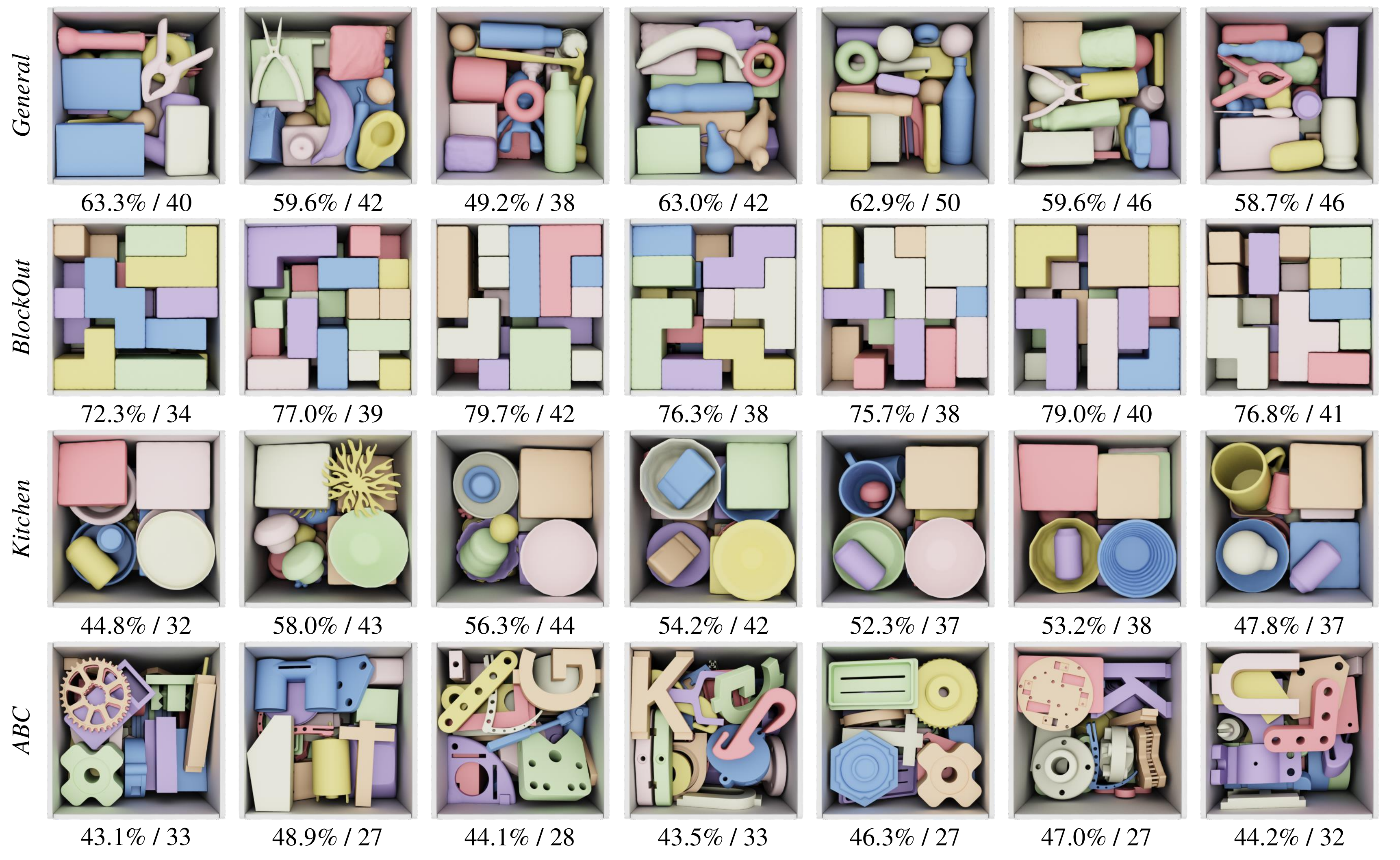}}
\vspace{-10pt}
\caption{Results generated by our buffered packing policies. These policies are trained with a buffer size $K=10$ and tested with $K=3$.}
\label{fig:galleryBuffer2}
\end{center}
\vskip -0.2in
\end{figure*}

\begin{figure*}[ht]
\begin{center}
\centerline{\includegraphics[width=0.92\textwidth]{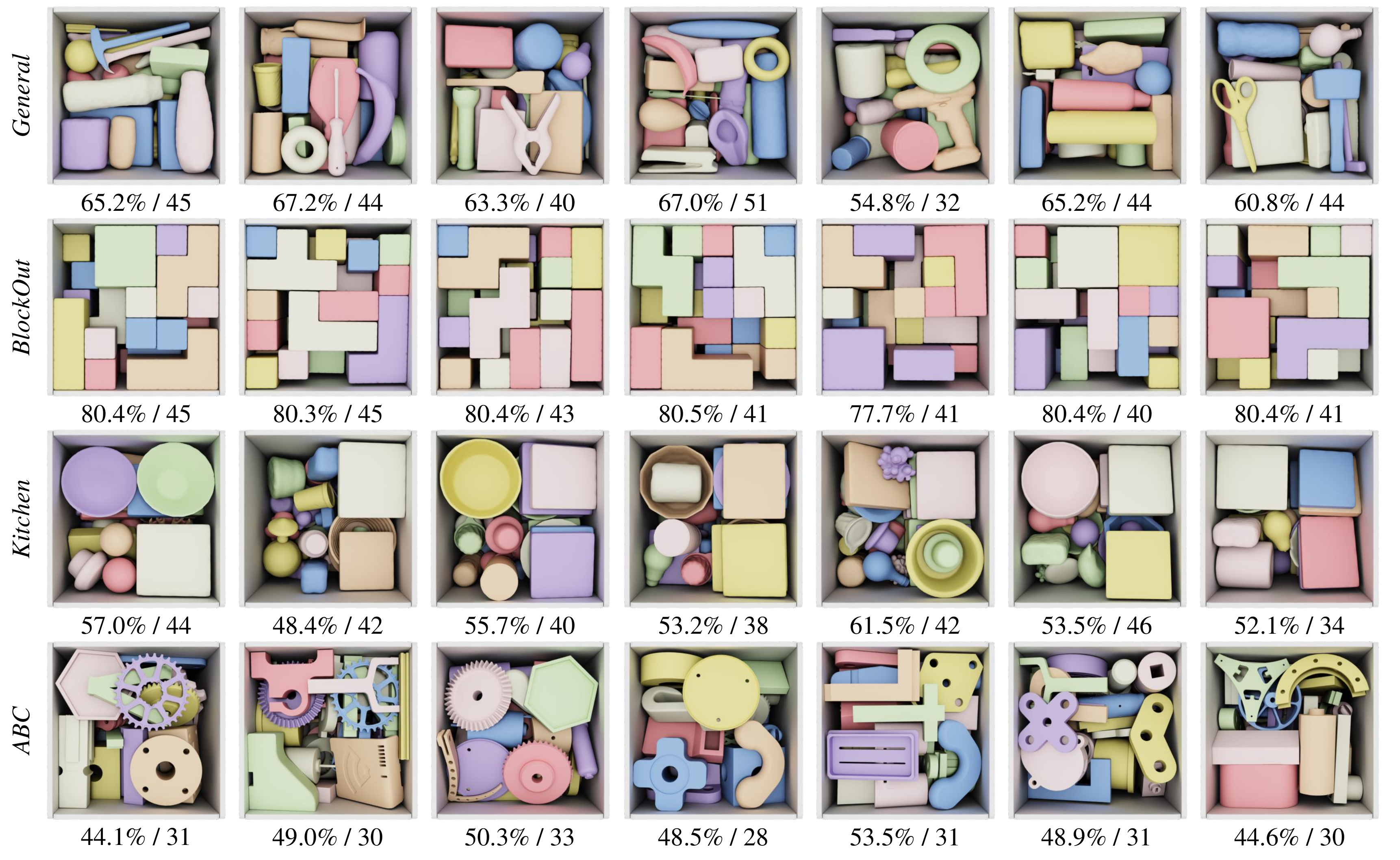}}
\vspace{-10pt}
\caption{Results generated by our buffered packing policies. These policies are trained with a buffer size $K=10$ and tested with $K=5$.}
\label{fig:galleryBuffer3}
\end{center}
\vskip -0.2in
\end{figure*}

We provide more visualization results tested on all datasets,  including  industrial shapes that come from the ABC dataset.
We show galleries of online PRP results on each dataset in~Fig.~\ref{fig:galleryOnline}. We provide qualitative results of buffered packing policies with $K=10$ in~Fig.~\ref{fig:galleryBuffer1}. The generalized results on buffered packing scenarios with $K=3$ and $K=5$ are visualized in~Fig.~\ref{fig:galleryBuffer2} and~Fig.~\ref{fig:galleryBuffer3}.

\end{document}